%% file: main.tex
\documentclass[journal]{IEEEtran}

\ifCLASSINFOpdf
   \usepackage[pdftex]{graphicx}
\else
   \usepackage[dvips]{graphicx}
\fi

\usepackage{hyperref}
\usepackage{amsmath}
\usepackage{subfig}
\usepackage{tikz}
\usepackage{xcolor}
\usepackage{multirow,booktabs}
\usepackage{float}
\usepackage{amsmath}                                                                        
\usepackage{amssymb} 
\usepackage{algorithm}
\usepackage{colortbl}
\usepackage{tabularx}
\usepackage{siunitx}
\usepackage{listing}
\usepackage[frozencache]{minted}
\usepackage[para]{footmisc}
\usepackage{color,soul}

\soulregister\cite7
\soulregister\ref7
\soulregister\SI7
\soulregister\circled7
\soulregister\videolink7
\soulregister\gitlink7

\newcolumntype{M}{>{\centering\arraybackslash}m{1cm}}

\usepackage[utf8]{inputenc}
\usepackage{algcompatible,amssymb}

\algnewcommand\algorithmicto{\textbf{to}}
\algnewcommand\RETURN{\State \textbf{return} }

\makeatletter
\newcommand\fs@norules{\def\@fs@cfont{\bfseries}\let\@fs@capt\floatc@ruled
  \def\@fs@pre{}%
  \def\@fs@post{}%
  \def\@fs@mid{\kern3pt}%
  \let\@fs@iftopcapt\iftrue}
\makeatother
\floatstyle{norules}
\restylefloat{algorithm}

\newcommand*\circled[1]{\tikz[baseline=(char.base)]{
            \node[shape=circle,fill,inner sep=1pt] (char) {\textcolor{white}{#1}};}}

\renewcommand{\baselinestretch}{.988}

\graphicspath{{./Figures/}}

\begin{document}

\title{A 64mW DNN-based Visual Navigation Engine\\for Autonomous Nano-Drones}

\author{Daniele~Palossi,
        ~Antonio~Loquercio,
        ~Francesco~Conti,
        ~\IEEEmembership{Member,~IEEE,}%
		~Eric~Flamand,
		~Davide~Scaramuzza,
		~\IEEEmembership{Member,~IEEE,}%
		~Luca~Benini,
		~\IEEEmembership{Fellow,~IEEE}%
\thanks{This work has been partially funded by projects EC H2020 OPRECOMP (732631) and ALOHA (780788), by the Swiss National Center of Competence Research (NCCR) Robotics and by the SNSF-ERC starting grant.}%
\thanks{D. Palossi, F. Conti, E. Flamand and L. Benini are with the Integrated System Laboratory of ETH Z\"urich, ETZ, Gloriastrasse 35, 8092 Z\"urich, Switzerland (e-mail: dpalossi@iis.ee.ethz.ch, fconti@iis.ee.ethz.ch, eflamand@iis.ee.ethz.ch, lbenini@iis.ee.ethz.ch).}%
\thanks{A. Loquercio and D. Scaramuzza are with the Robotic and Perception
Group, at both the Dep. of Informatics (University of Z\"urich) and the Dep. of Neuroinformatics (University of Z\"urich and ETH Z\"urich), Andreasstrasse 15, 8050 Z\"{u}rich, Switzerland.}%
\thanks{F. Conti and L. Benini are also with the Department of Electrical, Electronic and Information Engineering of University of Bologna, Viale del Risorgimento 2, 40136 Bologna, Italy (e-mail: f.conti@unibo.it, luca.benini@unibo.it).}%
\thanks{E. Flamand is also with GreenWaves Technologies, P\'{e}pini\`{e}re Berg\`{e}s, avenue des Papeteries, 38190 Villard-Bonnot, France (e-mail: eric.flamand@greenwaves-technologies.com).}
}

%


\newcommand{\efficientfps}[0]{6\,fps}
\newcommand{\efficientpowerboard}[0]{\SI{64}{\milli\watt}}
\newcommand{\efficientpoweravg}[0]{\SI{45}{\milli\watt}}
\newcommand{\efficientpowerpercent}[0]{0.8\%}
\newcommand{\efficientVDD}[0]{\SI{1.0}{V}}
\newcommand{\efficientFCfmax}[0]{\SI{50}{MHz}}
\newcommand{\efficientCLfmax}[0]{\SI{100}{MHz}}
\newcommand{\efficientlayerpowermax}[0]{\SI{47}{mW}}
\newcommand{\efficientlayerpowermin}[0]{\SI{13}{mW}}
\newcommand{\efficientlayerpoweravg}[0]{\SI{39}{mW}}

\newcommand{\fastfps}[0]{18\,fps}
\newcommand{\fastpowerboard}[0]{\SI{284}{\milli\watt}}
\newcommand{\fastpoweravg}[0]{\SI{272}{\milli\watt}}
\newcommand{\fastpowerpercent}[0]{3.5\%}
\newcommand{\fastVDD}[0]{\SI{1.2}{V}}
\newcommand{\fastFCfmax}[0]{\SI{250}{MHz}}
\newcommand{\fastCLfmax}[0]{\SI{250}{MHz}}

\newcommand{\videolink}{\url{https://youtu.be/57Vy5cSvnaA}}
\newcommand{\gitlink}{\url{https://github.com/pulp-platform/pulp-dronet}}

\maketitle

\begin{abstract}
Fully-autonomous miniaturized robots (e.g., drones), with artificial intelligence (AI) based visual navigation capabilities, are extremely challenging drivers of Internet-of-Things edge intelligence capabilities. 
Visual navigation based on AI approaches, such as deep neural networks (DNNs) are becoming pervasive for standard-size drones, but are considered out of reach for nano-drones with a size of a few \SI{}{\cm\squared}.
In this work, we present the first (to the best of our knowledge) demonstration of a navigation engine for autonomous nano-drones capable of closed-loop end-to-end DNN-based visual navigation. 
To achieve this goal we developed a complete methodology for parallel execution of complex DNNs directly on board  resource-constrained milliwatt-scale nodes. 
Our system is based on GAP8, a novel parallel ultra-low-power computing platform, and a \SI{27}{\gram} commercial, open-source CrazyFlie 2.0 nano-quadrotor.
As part of our general methodology, we discuss the software mapping techniques that enable the state-of-the-art deep convolutional neural network presented in~\cite{dronet} to be fully executed aboard within a strict \efficientfps{} real-time constraint with no compromise in terms of flight results, while all processing is done with only \efficientpowerboard{} on average.
Our navigation engine is flexible and can be used to span a wide performance range: at its peak performance corner, it achieves \fastfps{} while still consuming on average just \fastpowerpercent{} of the power envelope of the deployed nano-aircraft.
To share our key findings with the embedded and robotics communities and foster further developments in autonomous nano-UAVs, we publicly release all our code, datasets, and trained networks.
\end{abstract}


%
%
\section*{Supplementary Material}
Supplementary video at: \videolink.
The project's code, datasets and trained models are available at: \gitlink.
\input{01-introduction.tex}
\input{02-related_work.tex}
\input{03-background.tex}
\input{04-implementation.tex}
\input{05-prototype.tex}
\input{06-results.tex}
\input{07-conclusion.tex}

\section*{Acknowledgments}
The authors thank Hanna M\"uller for her contribution in designing the \textit{PULP-Shield}, No\'e Brun for his support in making the camera-holder, and Frank K. G\"urkaynak for his assistance in making the supplementary videos.

\input{main.bbl}

\vspace{-0.5cm}
\begin{IEEEbiography}[{\includegraphics[width=1in,height=1.25in,clip,keepaspectratio]{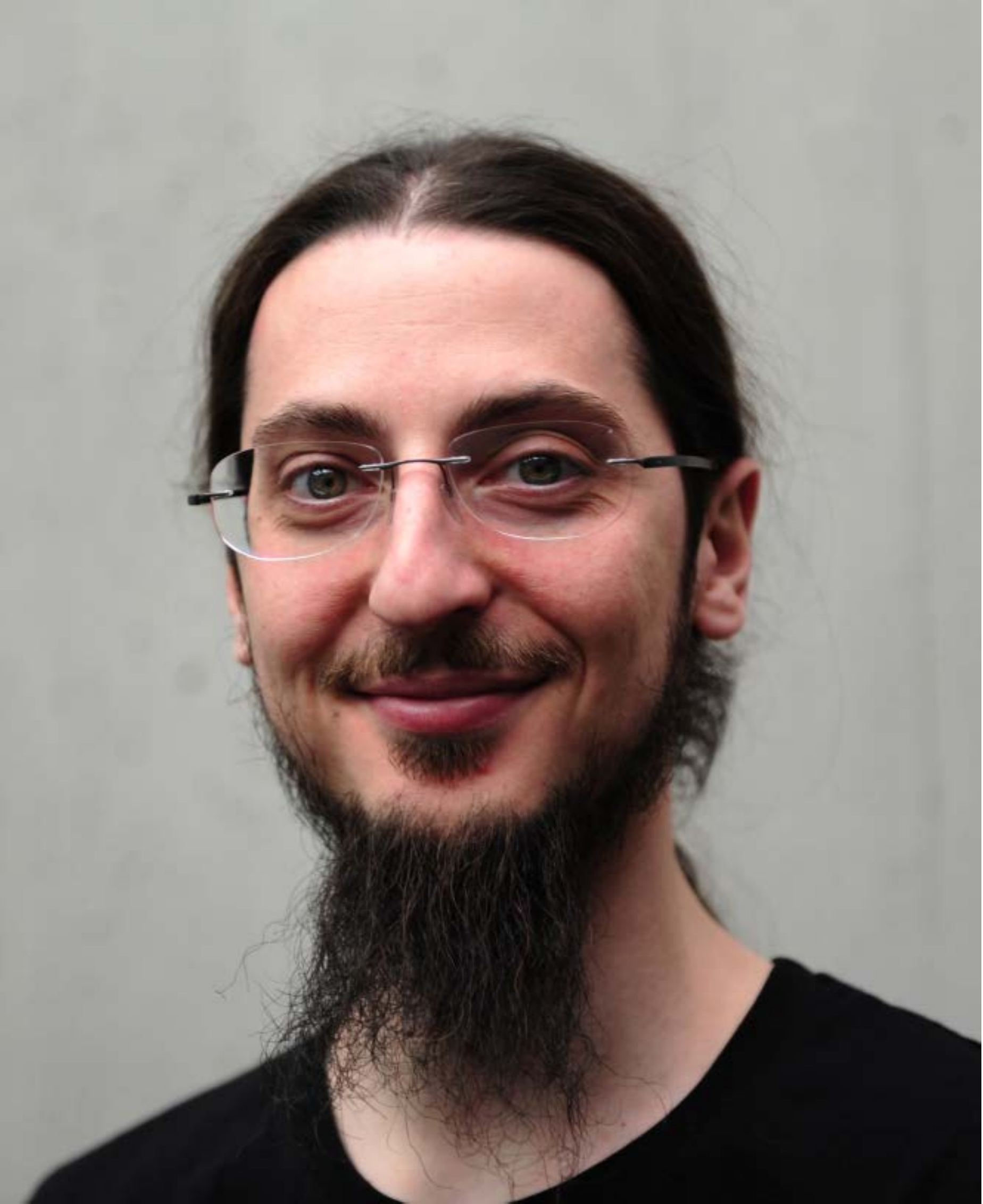}}]{Daniele Palossi} is a Ph.D. student at the Dept. of Information Technology and Electrical Engineering at the Swiss Federal Institute of Technology in Z\"{u}rich (ETH Z\"{u}rich).
He received his B.S. and M.S. in Computer Science Engineering from the University of Bologna, Italy.
In 2012 he spent six months as a research intern at ST Microelectronics, Agrate Brianza, Milano, working on 3D computer vision algorithms for the STM STHORM project.
In 2013 he won a one-year research grant at the University of Bologna, with a focus on design methodologies for high-performance embedded systems.
He is currently working on energy-efficient algorithms for autonomous vehicles and advanced driver assistance systems.
\end{IEEEbiography}

\begin{IEEEbiography}[{\includegraphics[width=1in,height=1.25in,clip,keepaspectratio]{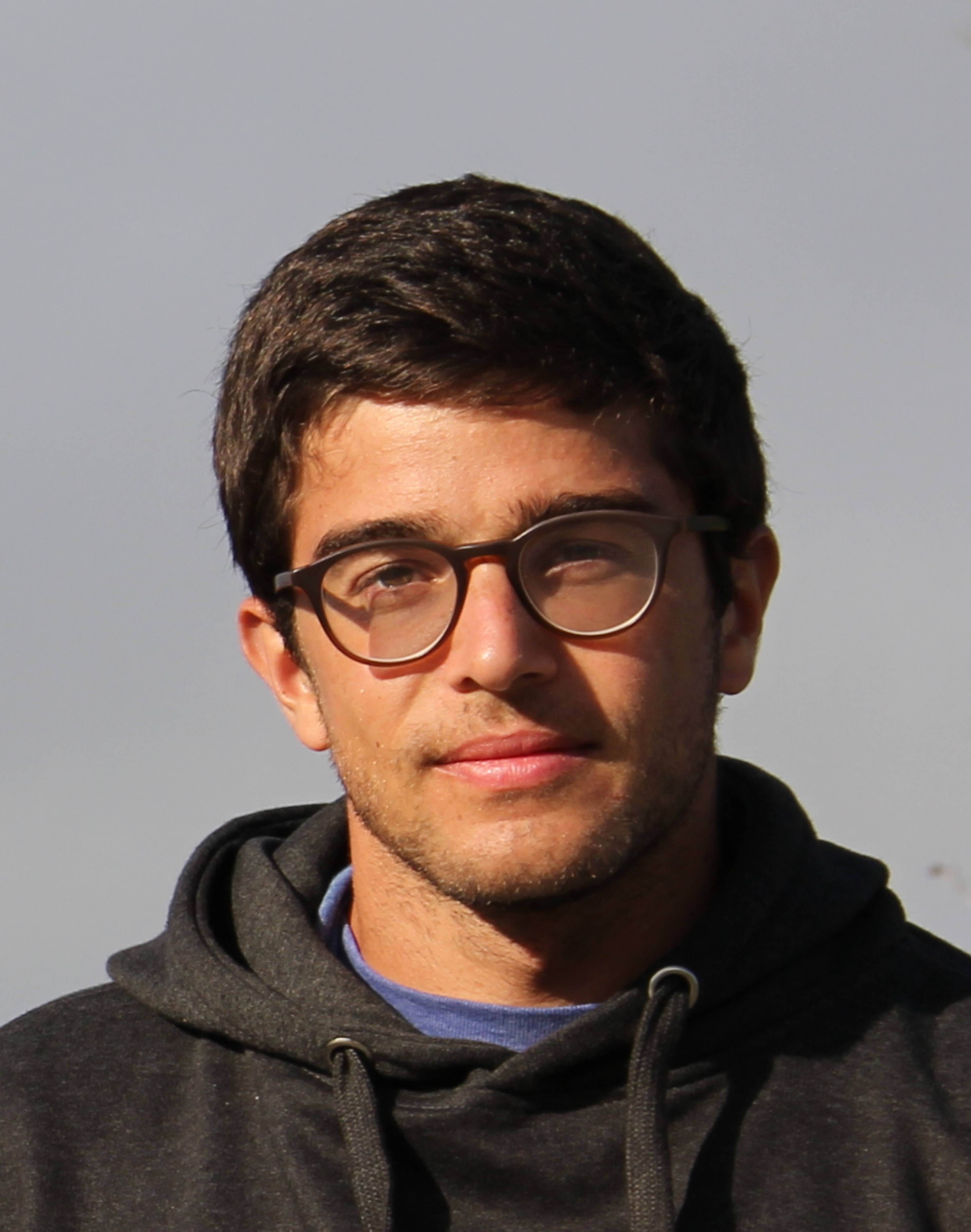}}]{Antonio Loquercio} received the MSc degree in Robotics, Systems and Control from ETH Z\"{u}rich in 2017.
He is working toward the Ph.D. degree in the Robotics and Perception Group at the University of Z\"{u}rich under the supervision of Prof. Davide Scaramuzza.
His main interests are data-driven methods for perception and control in robotics. 
He is a recipient of the ETH Medal for outstanding master thesis (2017).
\end{IEEEbiography}

\begin{IEEEbiography}[{\includegraphics[width=1in,height=1.25in,clip,keepaspectratio]{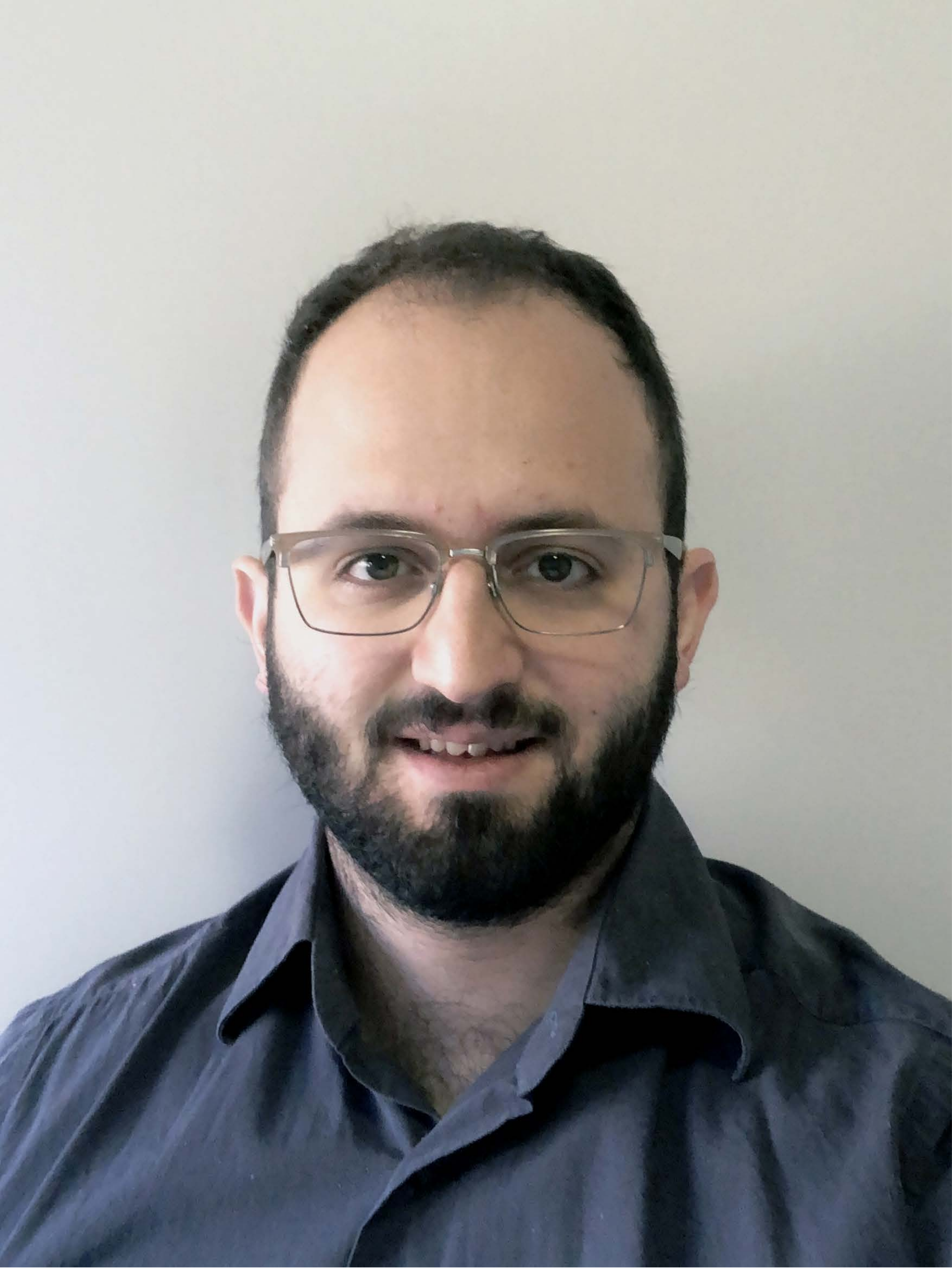}}]{Francesco Conti} received the Ph.D. degree from the University of Bologna in 2016 and is currently a post-doctoral researcher at the Integrated Systems Laboratory, ETH Z\"{u}rich, Switzerland, and the Energy-Efficient Embedded Systems Laboratory, University of Bologna, Italy.
His research focuses on energy-efficient multicore architectures and applications of deep learning to low power digital systems.
He has co-authored more than 30 papers in international conferences and journals, and he has been the recipient of three best paper awards (ASAP'14, EVW'14, ESWEEK'18) and the 2018 HiPEAC Tech Transfer Award.
\end{IEEEbiography}

\begin{IEEEbiography}[{\includegraphics[width=1in,height=1.25in,clip,keepaspectratio]{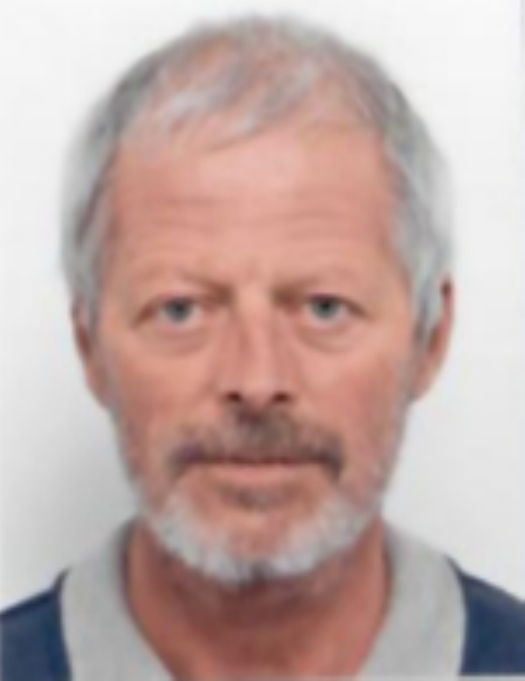}}]{Eric Flamand} got his Ph.D. in Computer Science from INPG, France, in 1982.
For the first part of his career, he worked as a researcher with CNET and CNRS in France, on architectural automatic synthesis, design and architecture, compiler infrastructure for highly constrained heterogeneous small parallel processors.
He then held different technical management in the semiconductor industry, first with Motorola where he was involved in the architecture definition and tooling of the StarCore DSP.
Then with ST Microelectronics first being in charge of all the software development of the Nomadik Application Processor and then in charge of the P2012 corporate initiative aiming at the development of a many-core device.
He is now co-founder and CTO of Greenwaves Technologies a French-based startup developing an IoT processor derived from the PULP project. 
He is also acting as a part-time research consultant for the ETH Z\"{u}rich.
\end{IEEEbiography}

\begin{IEEEbiography}[{\includegraphics[width=1in,height=1.25in,clip,keepaspectratio]{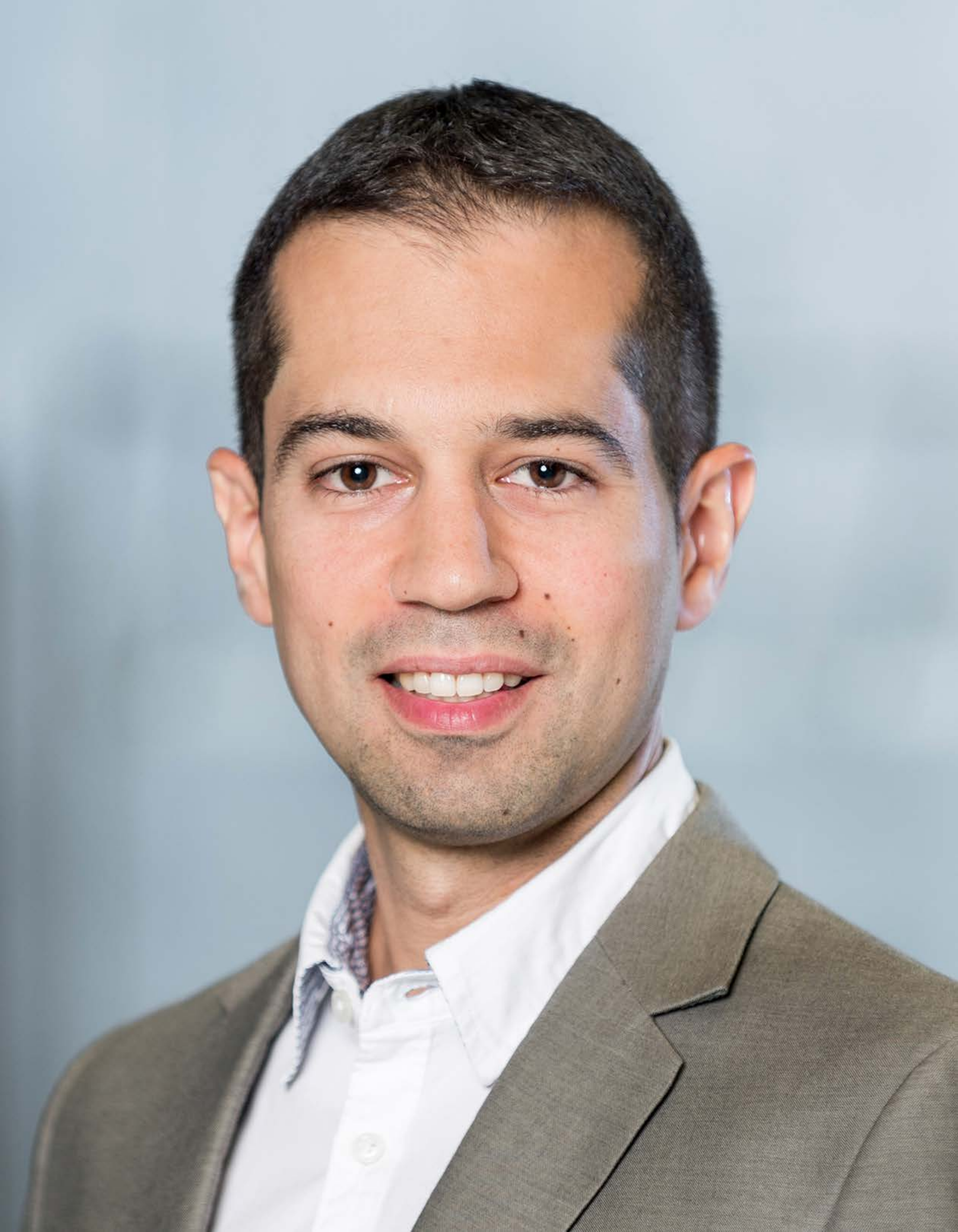}}]{Davide Scaramuzza} (1980, Italy) received the Ph.D. degree in robotics and computer vision from ETH Z\"{u}rich, Z\"{u}rich, Switzerland, in 2008, and a Postdoc at University of Pennsylvania, Philadelphia, PA, USA. He is a Professor of Robotics with University of Z\"{u}rich, where he does research at the intersection of robotics, computer vision, and neuroscience.
From 2009 to 2012, he led the European project sFly, which introduced the world's first autonomous navigation of microdrones in GPS-denied environments using visual-inertial sensors as the only sensor modality.
He coauthored the book Introduction to Autonomous Mobile Robots (MIT Press). 
Dr. Scaramuzza received an SNSF-ERC Starting Grant, the IEEE Robotics and Automation Early Career Award, and a Google Research Award for his research contributions.
\end{IEEEbiography}

\begin{IEEEbiography}[{\includegraphics[width=1in,height=1.25in,clip,keepaspectratio]{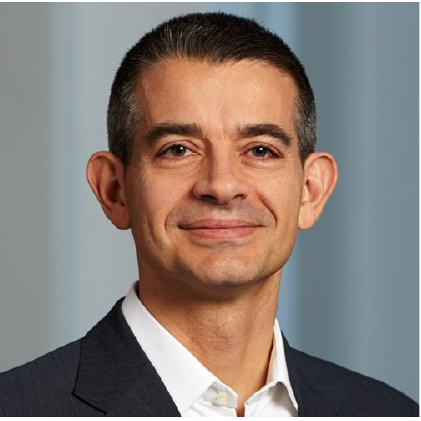}}]{Luca Benini} holds the chair of Digital Circuits and Systems at ETH Z\"{u}rich and is Full Professor at the Universit\`a di Bologna.
Dr. Benini's research interests are in energy-efficient system design for embedded and high-performance computing. 
He is also active in the area of energy-efficient smart sensors and ultra-low power VLSI design. 
He has published more than 1000 papers, five books and several book chapters. 
He is a Fellow of the IEEE and the ACM and a member of the Academia Europaea. 
He is the recipient of the 2016 IEEE CAS Mac Van Valkenburg Award.
\end{IEEEbiography}

\end{document}

%% file: 01-introduction.tex
\section{Introduction} \label{Sec:introduction}

\IEEEPARstart{W}{\lowercase{ith}} the rise of the Internet-of-Things (IoT) era and rapid development of artificial intelligence (AI), embedded  systems ad-hoc programmed to act in relative isolation are being progressively replaced by AI-based sensor nodes that acquire information, process and understand it, and use it to interact with the environment and with each other.
The "ultimate" IoT node will be capable of autonomously navigating the environment and, at the same time, sensing, analyzing, and understanding it~\cite{iotUAV_survey}.

\begin{figure}[t]
	\centering
	\includegraphics[width=\columnwidth]{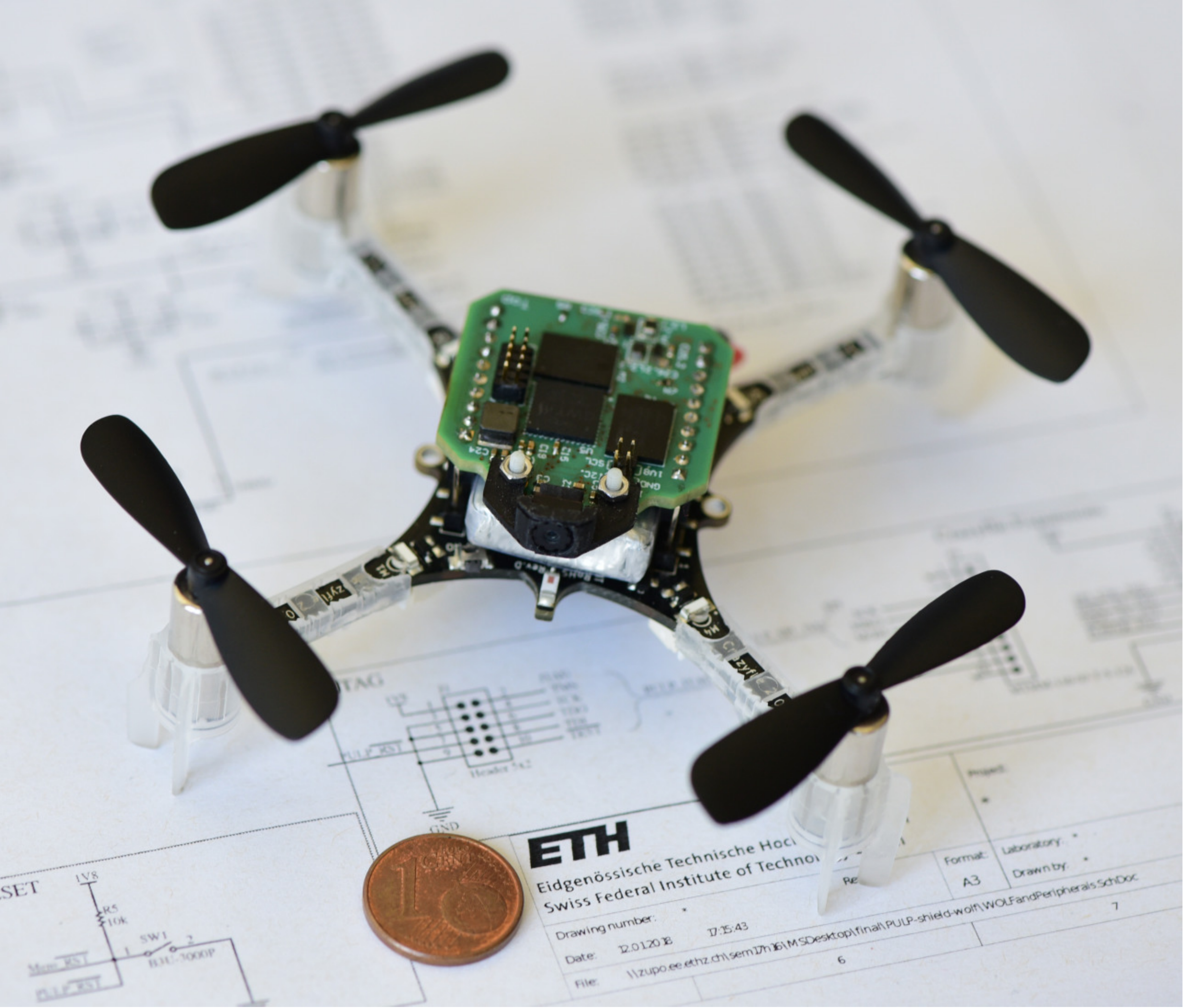}
	\caption{Our prototype based on the COTS \textit{Crazyflie 2.0} nano-quadrotor extended with our \textit{PULP-Shield}. The system can run the \textit{DroNet}~\cite{dronet} CNN for autonomous visual navigation up to \fastfps{} using only \textit{onboard} resources.}
	\label{fig:prototype}
\end{figure}

Fully autonomous nano-scale unmanned aerial vehicles (UAVs) are befitting embodiments for this class of smart sensors: with their speed and agility, they have the potential to quickly collect information from both their onboard sensors and from a plethora of devices deployed in the environment.
Nano-UAVs could also perform advanced onboard analytics, to pre-select essential information before transmitting it to centralized servers~\cite{ContiIoTEndpointSystemonChip2017}.
The tiny form-factor of nano-drones is ideal both for indoor applications where they should safely operate near humans (for surveillance, monitoring, ambient awareness, interaction with smart environments, etc.)~\cite{palossiJSPS18} and for highly-populated urban areas, where they can exploit complementary sense-act capabilities to interact with the surroundings (e.g., smart-building, smart-cities, etc.)~\cite{Floreano_2015}.
As an example, in this IoT scenario, a relevant application for intelligent nano-size UAVs can be the online detection of wireless activity, from edge nodes deployed in the environment, via onboard radio packet sniffing~\cite{HAWK14}. 

Commercial off-the-shelf (COTS) quadrotors have already started to enter the nano-scale, featuring only few centimeters in diameter and a few tens of grams in weight~\cite{Piccoli_2017}.
However, commercial nano-UAVs still lack the autonomy boasted by their larger counterparts~\cite{dronet, Lin_2017, 7989679, loianno2018special}, since their computational capabilities, heavily constrained by their tiny power envelopes, have been considered so far to be totally inadequate for the execution of sophisticated AI workloads, as summarized in Table~\ref{tab:taxonomy}.

\begin{table}[t]
\renewcommand{\arraystretch}{1.5}
\caption{Rotorcraft UAVs taxonomy by vehicle class-size.}
\label{tab:taxonomy}
\centering
\footnotesize
\resizebox{\columnwidth}{!}{
\begin{tabular}{|c|c|c|c|}
\hline
Vehicle Class & $\oslash$ : Weight [cm:kg] & Power [W] & Onboard Device\\
\hline
\text{\textit{std-size} \cite{iot_airQuality}}				& $\sim$ 50 : $\geq$ 1		& $\geq$ 100	& Desktop\\ 
\hline
\text{\textit{micro-size} \cite{conroy2009implementation}}	& $\sim$ 25 : $\sim$ 0.5 	& $\sim$ 50		& Embedded\\ 
\hline
\text{\textit{nano-size} \cite{40gSTM32OpticFlow}}			& $\sim$ 10 : $\sim$ 0.01 	& $\sim$ 5		& MCU\\ 
\hline
\text{\textit{pico-size} \cite{Wood2017}} 	                & $\sim$ 2 : $\leq$ 0.001	& $\sim$ 0.1 	& ULP\\ 
\hline
\end{tabular}
}
\end{table}

The traditional approach to autonomous navigation of a UAV is the so-called \textit{localization-mapping-planning} cycle, which consists of estimating the robot motion using either offboard (e.g., GPS~\cite{GPS_river}) or onboard sensors (e.g., visual-inertial sensors~\cite{Scaramuzza_2014}), building a local 3D map of the environment, and planning a safe trajectory through it~\cite{loianno2018special}. 
These methods, however, are very expensive for computationally-constrained platforms.
Recent results have shown that much lighter algorithms, based on convolutional neural networks (CNNs), are sufficient for enabling basic reactive navigation of small drones, even without a map of the environment~\cite{dronet, crash_fly, cad2rl, trailnet, dagger}.
However, their computational and power needs are unfortunately still above the allotted budget of current navigation engines of nano-drones, which are based on simple, low-power microcontroller units (MCUs). 

In Wood~et~al.~\cite{Wood2017}, the authors indicate that, for small-size UAVs, the maximum power budget that can be spent for onboard computation is 5\% of the total, the rest being used by the propellers (86\%) and the low-level control parts (9\%).
The problem of bringing state-of-the-art navigation capabilities on the challenging classes of nano- and pico-size UAVs is therefore strictly dependent on the development of energy-efficient and computationally capable hardware, highly optimized software and new classes of algorithms combined into a next-generation navigation engine.
These constraints and requirements depict the same scenario faced in deploying high-level computation capabilities on IoT edge-nodes/sensors.
Moreover, in the case of a flying miniature robot, the challenge is exacerbated by the strict real-time constraint dictated by the need for fast reaction time to prevent collisions with dynamic obstacles.

Whereas standard-size UAVs with a power envelope of several hundred Watts have always been able to host powerful high-end embedded computers like Qualcomm Snapdragon\footnote{https://developer.qualcomm.com/hardware/qualcomm-flight}, Odroid, NVIDIA Jetson TX1, and TX2, etc., most nano-sized UAVs have been constrained by the capabilities of microcontroller devices capable of providing a few hundred Mop/s at best.
Therefore, CNN-based autonomous vision navigation was so far considered to be out of reach for this class of drones.

In this work, we propose a novel visual navigation engine and a general methodology to deploy complex CNN on top of COTS resources-constrained computational edge-nodes such as a nano-size flying robot. 
We present what, to the best of our knowledge, is the first deployment of a State-of-the-Art (SoA), fully autonomous vision-based navigation system based on deep learning on top of a UAV visual navigation engine consuming less than \fastpowerboard{} at peak (\efficientpowerboard{} in the most energy-efficient configuration), fully integrated and in closed-loop control within an open source COTS \textit{CrazyFlie 2.0} nano-UAV.
Our visual navigation engine, shown on the top of the \textit{CrazyFlie 2.0} in Figure~\ref{fig:prototype}, leverages the \textit{GreenWaves Technologies GAP8} SoC, a high-efficiency embedded processor taking advantage of the emerging parallel ultra-low-power (PULP) computing paradigm to enable the execution of complex algorithmic flows onto power-constrained devices, such as nano-scale UAVs.

This work provides several contributions beyond the SoA of nano-scale UAVs and serves as a proof-of-concept for a broader class of AI-based applications in the IoT field.
In this work:
\begin{itemize}
\item we developed a general methodology for deploying SoA deep learning algorithms on top of ultra-low power embedded computation nodes, as well as a miniaturized robot;
\item we adapted \textit{DroNet}, the CNN-based approach for autonomous navigation proposed in Loquercio~et~al.~\cite{dronet} for standard-sized UAVs, to the computational requirements of a nano-sized UAV, such as fixed-point computation;
\item we deployed DroNet on the \textit{PULP-Shield}, an ultra-low power visual navigation module featuring the GAP8 SoC, an ultra-low power camera and off-chip Flash/DRAM memory; the shield is designed as a pluggable PCB for the \SI{27}{\gram} COTS \textit{CrazyFlie 2.0} nano-UAV;
\item we demonstrate our methodology for the DroNet CNN, achieving a comparable quality of results in terms of UAV control with respect to the standard-sized baseline of~\cite{dronet} within an overall
PULP-Shield power budget of just \efficientpowerboard{}, delivering a throughput of \efficientfps{} and up to \fastfps{} within \fastpowerboard{};
\item we field-prove our methodology presenting a closed-loop fully working demonstration of vision-driven autonomous navigation relying only on onboard resources.
\end{itemize}
Our work demonstrates that parallel ultra-low-power computing is a viable solution to deploy autonomous navigation capabilities on board nano-UAVs used as smart, mobile IoT end-nodes, while at the same time showcasing a complete hardware/software methodology to implement such complex workloads on a heavily power- and memory-constrained device.
We prove in the field the efficacy of our methodology by presenting a closed-loop fully functional demonstrator in the supplementary video material.
To foster further research on this field, we release the PULP-Shield design and all code running on GAP8, as well as datasets and trained networks, as publicly available under liberal open-source licenses.

The rest of the paper is organized as follows: Section~\ref{Sec:related_work} provides the SoA overview both in term of nano-UAVs and low-power IoT.
Section~\ref{Sec:background} introduces the software/hardware background of our work.
Section~\ref{Sec:implementation} presents in detail our CNN mapping methodology, including software tools and optimizations.
Section~\ref{Sec:prototype} discusses the design of the visual navigation engine.
Section~\ref{Sec:discussion} shows the experimental evaluation of the work, considering both performance and power consumption, comparing our results with the SoA and also evaluating the final control accuracy. 
Finally, Section~\ref{Sec:conclusion} concludes the paper.

%% file: 02-related_work.tex
\section{Related Work} \label{Sec:related_work}

The development of the IoT is fueling a trend toward edge computing, improving scalability, robustness, and security \cite{iotUAV_survey}. 
While today's IoT edge nodes are usually stationary, autonomous nano-UAVs can be seen as perfect examples of next-generation IoT end-nodes, with high mobility and requiring an unprecedented level of onboard intelligence.
The goal of this work is to make SoA visual autonomous navigation compatible with ultra-low power nano-drones, unlocking their deployment for IoT applications.
Therefore, this section focuses on related work on nano-aircrafts~\cite{Wood2017} and the deployment of DNN on top of low-power IoT nodes.

The traditional approach to autonomous navigation of nano-drones requires to offload computation to some remote, powerful base-station.
For instance, the authors of~\cite{dunkley14iros} developed a visual-inertial simultaneous localization and mapping (SLAM) algorithm, for a \SI{25}{\gram} nano quadrotor.
The SLAM algorithm was used to stabilize the robot and follow a reference trajectory.
All the computation was performed off-board, by streaming video and inertial information from the drone to a remote, power-unconstrained laptop.
The main problems with this class of solutions are latency, maximum communication distance, reliability issues due to channel noise, and high onboard power-consumption due to the high-frequency video streaming.

Few previous works presented nano-size flying robots with some degree of autonomous navigation relying on onboard computation.
In~\cite{40gSTM32OpticFlow}, the authors developed a \SI{4}{\gram} stereo-camera and proposed a velocity estimation algorithm able to run on the MCU on board a \SI{40}{\gram} flying robot.
If on one side this solution allows the drone to avoid obstacles during the flight, it still requires favorable flight conditions (e.g., low flight speed of \SI{0.3}{\meter/\second}).
In~\cite{briod2013optic}, an optical-flow-based guidance system was developed for a \SI{46}{\gram} nano-size UAV.
The proposed ego-motion estimation algorithm did not rely on feature tracking, making it possible to run on the onboard MCU.
Unfortunately, the autonomous functionality was limited to hovering, and the method did not reach the accuracy of computationally expensive techniques based on feature tracking.

In~\cite{navion}, an application-specific integrated circuit (ASIC), called \textit{NAVION}, for onboard visual-inertial odometry was presented.
Although this chip exposes enough computational power to perform state estimation up to \SI{171}{fps} within \SI{24}{\milli\watt}, this represents only one among other basic functionalities required by any UAV to be fully autonomous.
Therefore, in a real use case, the proposed ASIC would still need to be paired with additional circuits, both for complementary onboard computation as well as for interacting with the drone's sensors.
Moreover, to the date, the \textit{NAVION} accelerator does not reach the same level of maturity and completeness of our work; in fact, \textit{NAVION} has not yet been demonstrated on a real-life flying nano-drone. 

COTS nano-size quadrotors, like the \textit{Bitcraze Crazyflie 2.0} or the \textit{Walkera QR LadyBug}, embed on board low-power single core MCUs, like the \textit{ST Microelectronics} STM32F4~\cite{40gSTM32OpticFlow,dunkley14iros,7487496}.
While significant work has been done within academia~\cite{HegdeCaffePressooptimizedlibrary2016,KangCGOODCcodeGeneration2018,LokhmotovMultiobjectiveAutotuningMobileNets2018} and industry (e.g., TensorFlow Lite\footnote{ttps://www.tensorflow.org/lite\hspace{1cm}} and ARM Compute Library\footnote{https://arm-software.github.io/ComputeLibrary}) to ease the embedding of deep neural networks on mobile ARM-based SoC's, there is no general consensus yet on how to ``correctly'' deploy complex AI-powered algorithms, such a deep neural networks, on this class of low-power microcontrollers.
This is a ``hard'' problem both in terms of resource management (in particular available working memory and storage) and the peak throughput achievable by single core MCUs.
This problem is furthermore exacerbated by a lack of abstraction layers and computing facilities that are taken from granted by common deep learning tools, such as linear algebra libraries (e.g., BLAS, CUBLAS, CUDNN) and preprocessing libraries (e.g., OpenCV).

ARM has recently released CMSIS-NN~\cite{LaiCMSISNNEfficientNeural2018}, which is meant to shrink this gap by accelerating deep inference compute kernels on Cortex-M microcontroller platforms, providing the equivalent of a BLAS/CUDNN library (in Section~\ref{Sec:discussion} we present a detailed SoA comparison between our results and CMSIS-NN).
However, this effort does not curtail the difficulty of effectively deploying DNNs in memory-scarce platforms, which often requires particular scheduling/tiling~\cite{PeemenMemorycentricacceleratordesign2013,ZhangOptimizingFPGAbasedAccelerator2015} and is still widely considered an open problem.

Pushing beyond the aforementioned approaches, in this work we propose and demonstrate a visual navigation engine capable of sophisticated workloads, such as real-time CNN-based autonomous visual navigation~\cite{dronet}, entirely aboard within the limited power envelope of nano-scale UAVs ($\sim$\SI{0.2}{\watt}). 
Such a kind of autonomous navigation functionality has been previously limited to standard-sized UAVs, generally equipped with power-hungry processors ($\ge$10W) or relying on external processing and sensing (e.g., GPS)~\cite{HAWK14}. Our system relaxes both requirements: we use an onboard ultra-low-power processor and a learning-based navigation approach.

%% file: 03-background.tex
\section{Background} \label{Sec:background}

In this section, we summarize the hardware/software background underlying our visual navigation engine.
We first present the original \textit{DroNet} CNN developed for standard-size UAVs.
Then, we introduce the GAP8 SoC used on board of our nano-drone.

\begin{figure*}[t]
	\centering
	\includegraphics[width=\textwidth]{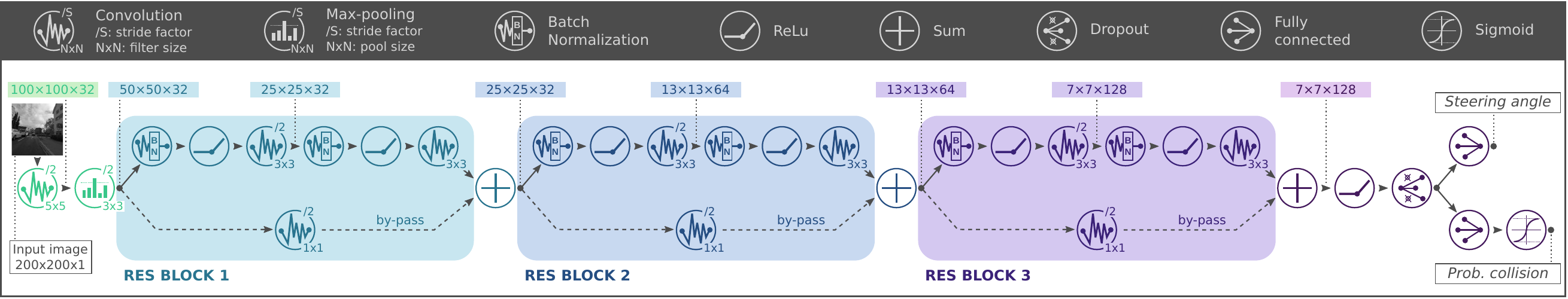}
	\caption{\textit{DroNet}~\cite{dronet} topology.}
	\label{fig:dronet-archi}
\end{figure*}

\subsection{DroNet}

DroNet is a lightweight residual CNN architecture.
By predicting the steering angle and the collision probability, it enables safe autonomous flight of a quadrotor in various indoor and outdoor environments.

The DroNet topology, as illustrated in Figure~\ref{fig:dronet-archi}, was inspired by residual networks~\cite{resnet} and was reduced in size to minimize the bare image processing time (inference).
The two tasks of steering and collision probability prediction share all the residual layers to reduce the network complexity and the frame processing time.
Then, two separate fully connected layers independently infer steering and collision probabilities.
Mean-squared error (MSE) and binary cross-entropy (BCE) have been used to train the two predictions, respectively.
A temporal dependent weighting of the two losses ensures the training convergence despite the different gradients' magnitude produced by each loss.
Eventually, to make the optimization focus on the samples that are most difficult to learn, hard negative mining was deployed in the final stages of learning.
The two tasks learn from two separate datasets.
Steering angle prediction was trained with the \textit{Udacity} dataset\footnote{https://www.udacity.com/self-driving-car}, while the collision probability was trained with the \textit{Z\"{u}rich bicycle} dataset\footnote{http://rpg.ifi.uzh.ch/dronet.html}.

The outputs of DroNet are used to command the UAV to move on a plane with velocity in forwarding direction $v_{k}$ and steering angle $\theta_{k}$.
More specifically, the low-pass filtered probability of collision is used to modulate the UAV forward velocity, while the low-pass filtered steering angle is converted to the drone's yaw control. 
The result is a single relatively shallow network that processes all visual information and directly produces control commands for a flying drone.
Learning the coupling between perception and control end-to-end provides several advantages, such as a simple, lightweight system and high generalization abilities.
Indeed, the method was shown to function not only in urban environments but also on a set of new application spaces without any initial knowledge about them~\cite{dronet}. 
More specifically, even without a map of the environment, the approach generalizes very well to scenarios completely unseen at training time, including indoor corridors, parking lots, and high altitudes.

\subsection{GAP8 Architecture} \label{Sec:background_pulp}

Our deployment target for the bulk of the DroNet computation is GAP8, a commercial embedded RISC-V multi-core processor derived from the PULP open source project\footnote{http://pulp-platform.org}.
At its heart, GAP8 is composed by an advanced RISC-V microcontroller unit coupled with a programmable octa-core accelerator with RISC-V cores enhanced for digital signal processing and embedded deep inference.

\begin{figure}[h]
	\centering
	\includegraphics[width=\columnwidth]{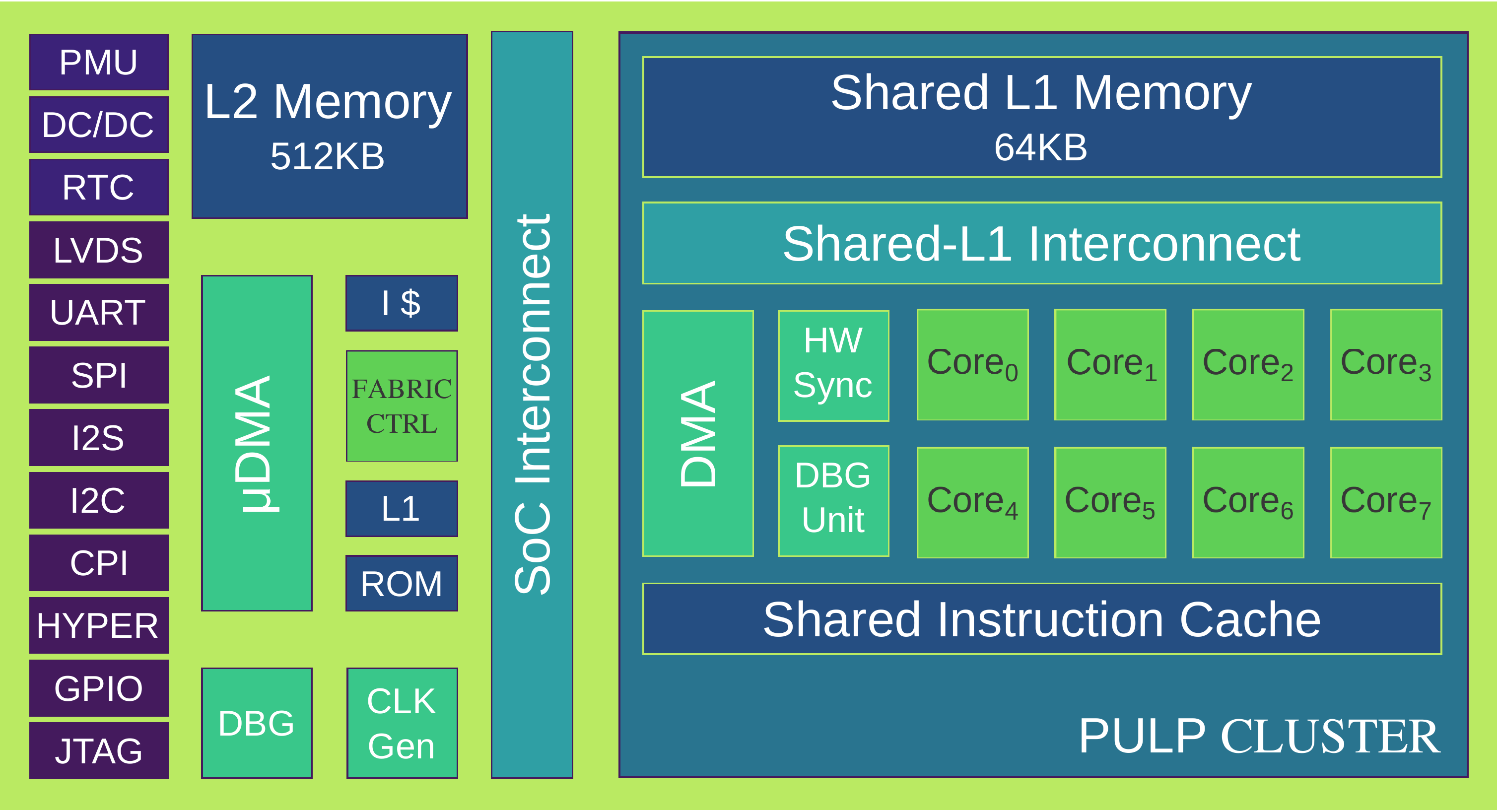}
	\caption{Architecture of the GAP8 embedded processor.}
	\label{fig:gap8_archi}
\end{figure}

Figure~\ref{fig:gap8_archi} shows the architecture of GAP8 in detail.
The processor is composed of two separate power and clock domains, the \textsc{fabric ctrl} (FC) and the \textsc{cluster} (CL).
The FC is an advanced microcontroller unit featuring a single RISC-V core coupled with \SI{512}{\kilo\byte} of SRAM (\textit{L2 memory}).
The FC uses an in-order, DSP-extended four-stage microarchitecture implementing the RISC-V instruction set architecture~\cite{GautschinearthresholdRISCVcore2016}.
The core supports the RV32IMC instruction set consisting of the standard ALU instructions plus the multiply instruction, with the possibility to execute compressed code.
In addition to this, the core is extended to include a register-register multiply-accumulate instruction, packed SIMD (single instruction multiple-data) DSP instructions (e.g., fixed-point dot product), bit manipulation instructions and two hardware loops.
Moreover, the SoC features an autonomous multi-channel I/O DMA controller ($\mu DMA$)~\cite{PulliniuDMAautonomoussubsystem2017} capable of transferring data between a rich set of peripherals (QSPI, I2S, I2C, HyperBus, Camera Parallel Interface) and the L2 memory with no involvement of the FC.
The HyperBus and QSPI interfaces can be used to connect GAP8 with an external DRAM or Flash memory, effectively extending the memory hierarchy with an external L3 having a bandwidth of \SI{333}{\mega\byte}/s and capacity up to \SI{128}{\mega\bit}.
Finally, the GAP8 SoC also includes a DC/DC converter converting the battery voltage down to the required operating voltage directly on-chip, as well as two separate frequency-locked loops (FLL) for ultra-low power clock generation~\cite{bellasi_fll}.

\begin{table*}[t]
\renewcommand{\arraystretch}{1.3}
\caption{DroNet accuracy on PULP. In \textbf{bold} the configuration used for the final deployment.}
\label{tab:dronet_pulp_accuracy}
\centering
\begin{tabular}{|c|c|c||c|c|c|c||c|c|}
\cline{1-9}
\multicolumn{3}{|c||}{Training}                                                                  & \multicolumn{6}{c|}{Inference - \texttt{Fixed16}} \\ 
\cline{1-9}
\multirow{2}{*}{Dataset}    & \multirow{2}{*}{Max-Pooling}  & \multirow{2}{*}{Data Type}    & \multicolumn{4}{c||}{Original Dataset}     & \multicolumn{2}{c|}{HiMax Dataset} \\
\cline{4-9}
                                &                               &                               & EVA   & RMSE  & Accuracy  & F1-score      & Accuracy  & F1-score \\
\cline{1-9}
\multirow{4}{*}{Original}       & $3\times3$                    & \texttt{Float32}              & 0.758 & 0.109 & 0.952     & 0.888         & 0.859     & 0.752 \\
\cline{2-9}
                                &   $3\times3$                  & \texttt{Fixed16}              & 0.746 & 0.115 & 0.946     & 0.878         & 0.841     & 0.798 \\
\cline{2-9}
                                &   $2\times2$                  & \texttt{Float32}              & 0.766 & 0.105 & 0.945     & 0.875         & 0.845     & 0.712  \\
\cline{2-9}
                                &   $2\times2$                  & \texttt{Fixed16}              & 0.795 & 0.097 & 0.935     & 0.857         & 0.873     & 0.774  \\
\cline{1-9}
\multirow{4}{*}{\textbf{Original + HiMax}}       &   $3\times3$                  & \texttt{Float32}              & 0.764 & 0.104 & 0.949     & 0.889         & 0.927     & 0.884 \\
\cline{2-9}
                                &   $3\times3$                  & \texttt{Fixed16}              & 0.762 & 0.109 & 0.956     & 0.894         & 0.918     & 0.870 \\
\cline{2-9}
                                &   $2\times2$                  & \texttt{Float32}              & 0.747 & 0.109 & 0.964     & 0.916         & 0.900     & 0.831 \\
\cline{2-9}
                              &   $\boldsymbol{2\times2}$                 & \textbf{\texttt{Fixed16}}              & \textbf{0.732} & \textbf{0.110} & \textbf{0.977}     & \textbf{0.946}         & \textbf{0.891}    & \textbf{0.821}\\
\cline{1-9}
\end{tabular}
\end{table*}

The \textsc{cluster} is dedicated to the acceleration of computationally intensive tasks.
It contains eight RISC-V cores (identical to the one used in the FC) sharing a \SI{64}{\kilo\byte} multi-banked \textit{shared L1 scratchpad memory} through a low-latency, high-throughput logarithmic interconnect~\cite{rahimi_fullysynthesizable_2011}.
The shared L1 memory supports single-cycle concurrent access from different cores requesting memory locations on separate banks and a starvation-free protocol in case of bank contentions (typically $<$10\% on memory-intensive kernels).
The eight cores are fed with instruction streams from a single shared, multi-ported I-cache to maximize the energy efficiency on the data-parallel code.
A cluster \textit{DMA} controller is used to transfer data between the shared L1 scratchpad and the L2 memory; it is capable of 1D and 2D bulk memory transfer on the L2 side (only 1D on the L1 side).
A dedicated \textit{hardware synchronizer} is used to support fast event management and parallel thread dispatching/synchronization to enable ultra-fine grain parallelism on the cluster cores.
\textsc{Cluster} and \textsc{fabric ctrl} share a single address space and communicate with one another utilizing two 64-bit AXI ports, one per direction.
A software runtime resident in the FC overviews all tasks offloaded to the CL and the $\mu DMA$.
On a turn, a low-overhead runtime on the CL cores exploits the hardware synchronizer to implement shared-memory parallelism in the fashion of OpenMP~\cite{contiDATE16}.

%% file: 04-implementation.tex
\section{CNN Mapping Methodology} \label{Sec:implementation}

In this section, we discuss and characterize the main methodological aspects related to the deployment of \textit{DroNet} on top of the GAP8 embedded processor. 
This task showcases all the main challenges for a typical deep learning application running on resource-constrained embedded IoT node.
Therefore, while our visual navigation engine is application-specific, the underlying methodology we present in the following of this section is general and could also be applied to other resource-bound embedded systems where computationally intensive tasks have to be performed under a real-time constraint on a parallel architecture.

\subsection{Deploying DroNet on GAP8} \label{Sec:deployment}

Following an initial characterization phase, we calculated the original convolutional neural network (CNN) to involve $\sim$41~MMAC operations per frame (accounting only for convolutional layers) and more than \SI{1}{\mega\byte} of memory needed solely to store the network's weights, yielding a baseline for the number of resources required on our navigation engine\footnote{
The baseline MMAC count does not correspond to the final implementation's instruction count, because it does not account for implementation details such as data marshaling operations to feed the processing elements; however, it can be used to set an upper bound to the minimum execution performance that is necessary to deploy DroNet at a given target frame rate.}.
To successfully deploy the CNN on top of GAP8, the execution of DroNet has to fit within the strict real-time constraints dictated by the target application, while respecting the bounds imposed by the on-chip and onboard resources.
Specifically, these constraints can be resumed in three main points:
\begin{itemize}
    \item the \textit{minimum real-time frame-rate} required to select a new trajectory on-the-fly or to detect a suspected obstacle in time to prevent a potential collision;
    \item the native \textit{quality-of-results} must be maintained when using an embedded ultra-low power camera (in our prototype, the \textit{HiMax} -- see Section~\ref{Sec:prototype} for details) instead of the high-resolution camera used by the original DroNet;
    \item the \textit{amount of available memory} on the GAP8 SoC, as reported in Section~\ref{Sec:background_pulp} we can rely on \SI{512}{\kilo\byte} of L2 SRAM and \SI{64}{\kilo\byte} of shared L1 scratchpad (TCDM), sets an upper bound to the size of operating set and dictates ad-hoc memory management strategy.
\end{itemize}

Therefore, it is clear there is a strong need for a strategy aimed at reducing the memory footprint and computational load to more easily fit within the available resources while exploiting the architectural parallelism at best to meet the real-time constraint.
The original DroNet network~\cite{dronet} has been modified to ease its final deployment; we operated incrementally on the model and training flow provided by the original DroNet, based on Keras/TensorFlow\footnote{https://github.com/uzh-rpg/rpg\_public\_dronet\hspace{2.5cm}}.

The first change we performed is the reduction of the numerical representation of weights and activations from the native one, 32-bit floating point (\texttt{Float32}), down to a more economical and hardware-friendly 16-bit fixed point one (\texttt{Fixed16}) that is better suited for the deployment on any MUC-class processor without floating point unit (FPU), like in our GAP8 SoC.
By analyzing the native \texttt{Float32} network post-training, we determined that a dynamic range of $\pm 8$ is sufficient to represent all weights and intermediate activations with realistic inputs.
Accordingly, we selected a \texttt{Fixed16} Q4.12 representation, using 4 bits for the integer part (including sign) and 12 bits for the fractional part of both activations and weights (rounding down to a precision of $2^{-12}$).
Then, we retrained the network from scratch replacing the native convolutional layers from Keras to make them ``quantization-aware'', using the methodology proposed by Hubara~et~al.~\cite{HubaraQuantizedNeuralNetworks2016}.

The second significant change with respect to the original version of DroNet is the extension of the collision dataset used in~\cite{dronet} (named \textit{Original} dataset) with $\sim$1300 images (1122 for training and 228 for test/validation) acquired with the same camera that is available aboard the nano-drone (named \textit{HiMax} dataset).
Fine-tuning approaches, like dataset extension, have proved to be particularly effective at improving network generalization capability \cite{Razavian2014transfer}. 
In our case, the original dataset is built starting form high-resolution color cameras whose images are significantly different from the ones acquired by the ULP low-resolution grayscale camera available in our navigation engine, particularly in terms of contrast.
Therefore, we extended the training set and we evaluate our CNN for both datasets separately. 
Finally, we modified the receptive field of max-pooling layers from 3$\times$3 to 2$\times$2, which yields essentially the same final results while reducing the execution time of max-pooling layers by 2.2$\times$ and simplifying their final implementation on GAP8.

Table~\ref{tab:dronet_pulp_accuracy} summarizes the results in terms of accuracy for all these changes.
Explained variance\footnote{EVA = $\frac{Var[y_{true}-y_{pred}]}{Var[y_{true}]}$} (EVA) and root-mean-squared error (RMSE) refer to the regression problem (i.e., steering angle) whereas Accuracy and F1-score\footnote{F-1 = $2\frac{precision \times recall}{precision+recall}$} are related to the classification problem (i.e., collision probability), evaluated on both the \textit{Original} and \textit{HiMax} datasets.
Regarding the \textit{Original} dataset, it is clear that the proposed modifications are not penalizing the overall network's capabilities.
Moreover, fine-tuning increases performance for almost all cases (both regression and classification), considering the test on the \textit{HiMax} dataset, there is a definite improvement in term of collision accuracy when training is done with the extended dataset. 
If we consider paired configurations, the fine-tuned one based is always outperforming its counterpart, up to 8\% in accuracy (i.e., max-pooling 3$\times$3, \texttt{Fixed16}).
In Table~\ref{tab:dronet_pulp_accuracy} we also highlight (in bold) the scores achieved by the final version of DroNet deployed on GAP8.

\subsection{AutoTiler} \label{Sec:tiling}

One of the most significant constraints in ULP embedded SoC's without caches is the explicit management of the memory hierarchy; that is, how to marshal data between the bigger - and slower - memories and the smaller - but faster - ones tightly coupled to the processing elements.
A common technique is \textit{tiling}~\cite{tilingBenini17}, which involves \textit{i}) partitioning the input and output data spaces in portions or tiles small enough to fit within the smallest memory in the hierarchy (in our case, the shared L1) and \textit{ii}) setting up an outer loop iterating on tiles, with each iteration comprising the loading of an input tile into the L1, the production of an output tile, and the storage of the output tile into the higher levels of the memory hierarchy.
Tiling is particularly effective for algorithms like deep neural networks exposing very regular execution and data access patterns.
As part of this work, we propose a tiling methodology that optimizes memory utilization on GAP8, while at the same time relieving the user from tedious and error-prone manual coding of the tiling loop and of the data movement mechanism.

\begin{figure}[h]
	\centering
	\includegraphics[width=\columnwidth]{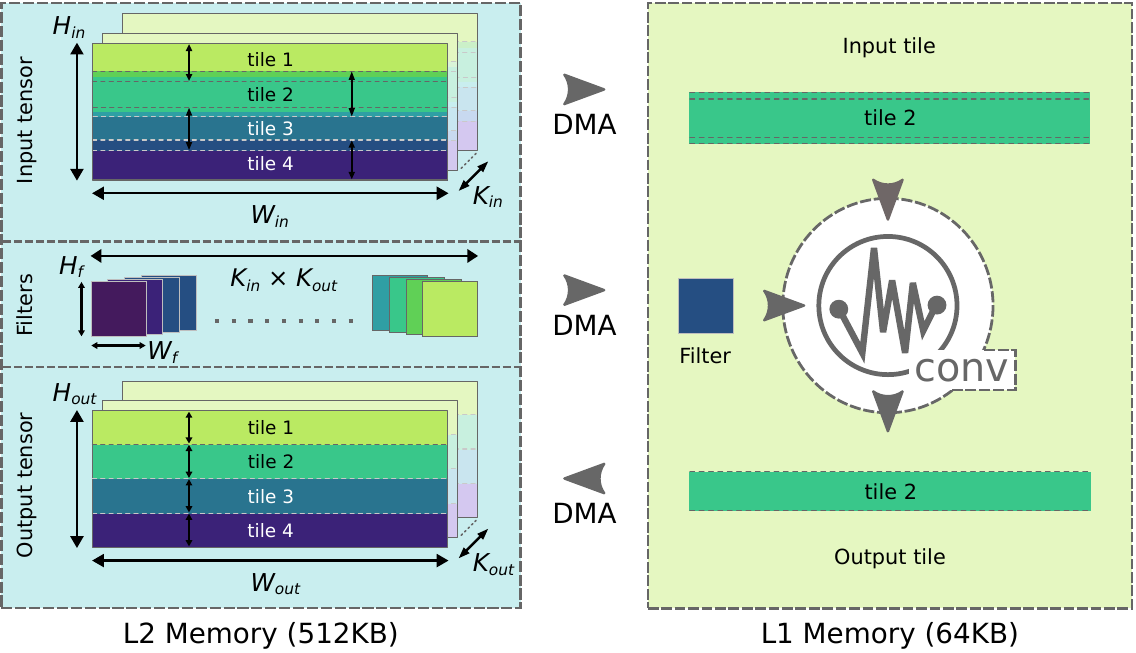}
	\caption{Convolutional layer tiling.}
	\label{fig:tiling}
\end{figure}

Considering Figure~\ref{fig:tiling} as a reference, each layer in a CNN operates on a three-dimensional input tensor representing a \textit{feature space} (with one \textit{feature map} per channel) and produces a new 3D tensor of activations as output.
Convolutional layers, in particular, are composed of a linear transformation that maps $K_{in}$ input feature maps into $K_{out}$ output feature maps employing of $K_{in}\times K_{out}$ convolutional filters (or weight matrices).
Therefore, in any convolutional layer, we can identify three different data spaces which can be partitioned in tiles in one or more of the three dimensions (i.e., $W$, $H$, and $K$ in Figure~\ref{fig:tiling}).
Similar considerations can also be made for the other layers in a CNN, allowing to treating them in the same fashion.

As the design space defined by all possible tiling schemes is very large, we developed a tool called \textit{AutoTiler} to help explore a subset of this space, choose an optimal tiling configuration, and produce C wrapping code that orchestrates the computation in a pipelined fashion as well as double-buffered memory transfers, taking advantage of the cluster DMA controller to efficiently move data between the L2 and L1 memories.
The fundamental unit of computation assumed by the \textit{AutoTiler} tool is the \textit{basic kernel}, a function considering that all its working data is already located in the L1 shared memory.
Examples of basic kernels include convolution, max-pooling, ReLU rectification, addition. 
To map the overall high-level algorithm to a set of basic kernels that operate iteratively on tiles, the \textit{AutoTiler} introduces a second level of abstraction: the \textit{node kernel}.
The structure of the target algorithm is coded by the developer as a dependency graph, where each node (a node kernel) is a composition of one or more basic kernels together with a specification of the related iteration space over $W$, $H$, $K_{in}$, $K_{out}$.
For example, a node kernel for a convolutional layer can be composed of a first basic kernel for setting the initial bias, a central one to perform convolutions and a final one for ReLU rectification: in the \textit{prologue}, \textit{body}, and \textit{epilogue}, respectively.
The \textit{AutoTiler} treats the tiling of each node kernel as an independent optimization problem constrained by the node kernel specification and the memory sizes.
This approach allows to build complex execution flows reusing hand-optimized \textit{basic kernels} and abstracting the underneath complexity from the developer.

\subsection{Tiling, Parallelization \& Optimization} \label{Sec:parallelization}

As introduced in Section~\ref{Sec:background}, the GAP8 SoC features 8+1 RISC-V cores with DSP-oriented extensions.
To develop an optimized, high-performance and energy-efficient application for GAP8 and meet the required real-time constraint it is paramount that the most computationally intensive kernels of the algorithm are parallelized to take advantage of the 8-core cluster and are entirely using the available specialized instructions.
For the purpose of this work, we used the \textit{AutoTiler} to fully implement the structure of the modified DroNet, therefore these optimization steps are reduced to hand-tuned parallelization and optimization of the basic kernels.

\begin{listing}[tb]
\begin{minted}[mathescape=true,
             escapeinside=||,
             numbersep=5pt,
             gobble=2,
             fontsize=\footnotesize,
             framesep=2mm]{python}
            
  # weight DMA-in
  DMA_Copy(|$\widehat{w} \leftarrow w$|)
  for |$t$| in range(nb_tiles_H): # tiling over $H$
    # prologue operation (set bias value)
    |$\widehat{y} \leftarrow$| BasicKernel_SetBias(|$\widehat{y}$|)
    for |$j$| in range(nb_tiles_Kin): # tiling over $K_{in}$
      # input tile DMA-in
      DMA_Copy(|$\widehat{x} \leftarrow x[j,t]$|)
      for |$i$| in range(|$K_{out}$|):
        # body operation (convolution)
        |$\widehat{y} \leftarrow$| BasicKernel_Conv_Spatial(|$\widehat{y}$|)
    |$\widehat{y} \leftarrow$| BasicKernel_ReLU(|$\widehat{y}$|)
    # output tile DMA-out
    DMA_Copy(|$y[i,t] \leftarrow \widehat{y}$|)
\end{minted}
\caption{Example of \textit{spatial} execution scheme. $x$, $w$, $y$ are the multi-dimensional input, weight and output tensors in L2 memory;  $\widehat{x}$, $\widehat{w}$, and $\widehat{y}$ are their respective tiles in L1 memory.} 
\label{lst:tiling_spatial}
\end{listing}
\begin{listing}[tb]
\begin{minted}[mathescape=true,
             escapeinside=||,
             numbersep=5pt,
             gobble=2,
             fontsize=\footnotesize,
             framesep=2mm]{python}

  for |$i$| in range(nb_tiles_Kout): # tiling over $K_{out}$
    # weight DMA-in
    DMA_Copy(|$\widehat{w} \leftarrow w[i]$|)
    # prologue operation (set bias value)
    |$\widehat{y} \leftarrow $|BasicKernel_SetBias(|$\widehat{y}$|)
    for |$j$| in range(nb_tiles_Kin): # tiling over $K_{in}$
      # input tile DMA-in
      DMA_Copy(|$\widehat{x} \leftarrow x[j]$|)
      # body operation (convolution)
      |$\widehat{y} \leftarrow$| BasicKernel_Conv_FeatWise(|$\widehat{w},\widehat{x},\widehat{y}$|)
    # epilogue operation (ReLU)
    |$\widehat{y} \leftarrow$| BasicKernel_ReLU(|$\widehat{y}$|)
    # output tile DMA-out
    DMA_Copy(|$y[i] \leftarrow \widehat{y}$|)
\end{minted}
\caption{Example of \textit{feature-wise} execution scheme. $x$, $w$, $y$ are the multi-dimensional input, weight and output tensors in L2 memory;  $\widehat{x}$, $\widehat{w}$, and $\widehat{y}$ are their respective tiles in L1 memory.} 
\label{lst:tiling_featurewise}
\end{listing}

\begin{figure}[tb]
	\centering
	\includegraphics[width=\columnwidth]{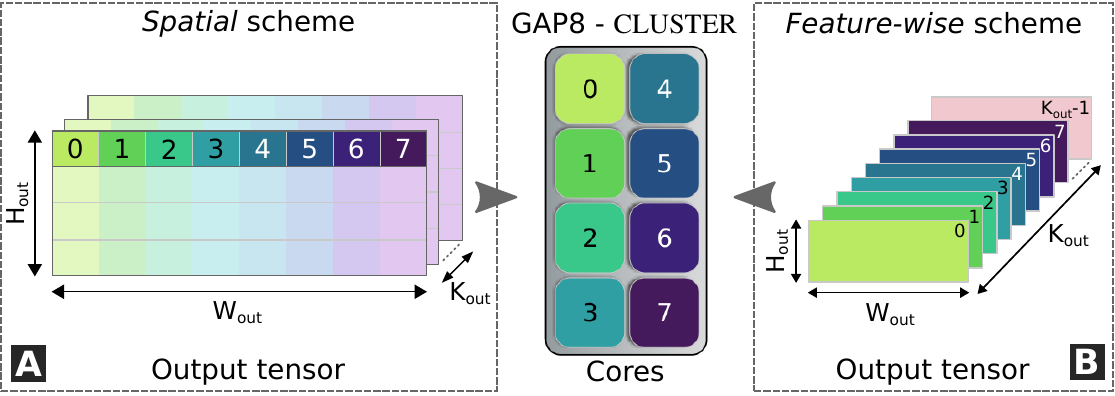}
	\caption{Parallelization schemes utilized in the \textit{DroNet} layers for deployment on GAP8; the different colors represent allocation to a different core.}
	\label{fig:conv_schemes}
\end{figure}

To exploit the available computational/memory resources at best, we constrain the \textit{AutoTiler} to target the following general scheme: the input tensor is tiled along the $H_{in}$ and $K_{in}$ dimensions, while the output tensor is tiled along $H_{out}$ and $K_{out}$ ones. 
The stripes along $H_{in}$ are partially overlapped with one another to take into account the receptive field of convolutional kernels at the tile border.
Execution of the node kernel happens in either a \textit{spatial} or \textit{feature-wise} fashion, which differ in the ordering of the tiling loops and in the parallelization scheme that is applied.
In the spatial scheme, work is split among parallel cores along the $W_{out}$ dimension; Figure~\ref{fig:tiling} and~\ref{fig:conv_schemes}-A refer to this scheme, which is also exemplified in Listing~\ref{lst:tiling_spatial}.
In the feature-wise scheme, which we only apply on full feature maps (i.e., the number of tiles in the $H_{out}$ direction is 1), work is split among cores along the $K_{out}$ dimension; this scheme is shown in Figure~\ref{fig:conv_schemes}-B and Listing~\ref{lst:tiling_featurewise}.
The choice of one scheme over the other is influenced mostly by the parallelization efficiency: after an exploration phase, we found the best performance arose when using the spatial scheme for the first node kernel of DroNet (first convolution + max-pooling) while using the feature-wise approach for the rest.
This choice is related to the fact that in deeper layers the feature map size drops rapidly and the spatial scheme becomes suboptimal because the width of each stripe turns too small to achieve full utilization of the cores.

To further optimize the DroNet execution, we made use of all the optimized signal processing instructions available in GAP8.
These include packed-SIMD instructions capable of exploiting sub-word parallelism, as well as bit-level manipulation and shuffling, which can be accessed by means of compiler intrinsics such as  {\footnotesize\texttt{\_\_builtin\_pulp\_dotsp2}} (for 16-bit dot product with 32-bit accumulation), {\footnotesize\texttt{\_\_builtin\_shuffle}} (permutation of elements within two input vectors), {\footnotesize\texttt{\_\_builtin\_pulp\_pack2}} (packing two scalars into a vector).

\subsection{L2 Memory Management Strategy} \label{Sec:implementation_memory}

\begin{figure}[t]
	\centering
	\includegraphics[width=\columnwidth]{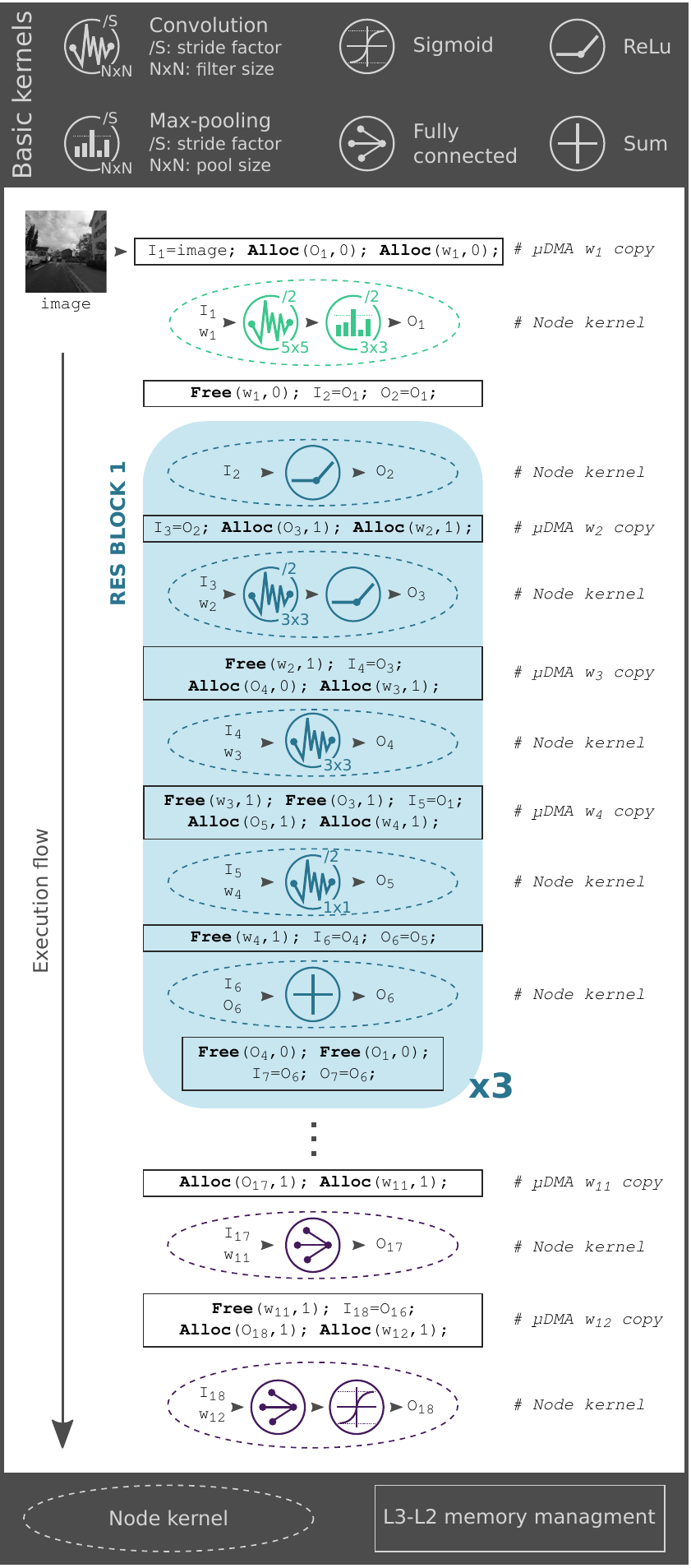}
	\caption{\textit{DroNet} on PULP execution graph (with pseudo-code).}
	\label{fig:exec_flow}
\end{figure}

Given \textit{i)} the residual-network topology of DroNet, which requires to increase the lifetime of the output tensors of some of the layers (due to bypass layers), and \textit{ii)} the ``scarcity'' of  L2 memory as a resource to store all weights and temporary feature maps (we would need more than \SI{1}{\mega\byte} in view of \SI{512}{\kilo\byte} available), an ad-hoc memory management strategy for the L2 memory is required, similar to what is done between L2 and L1 using the GAP8 \textit{AutoTiler}.
Due to the high energy cost of data transfers between L3 and L2, the strategy needs to be aimed at the maximization of the L2 reuse.

\begin{figure}[t]
	\centering
	\includegraphics[width=0.9\columnwidth]{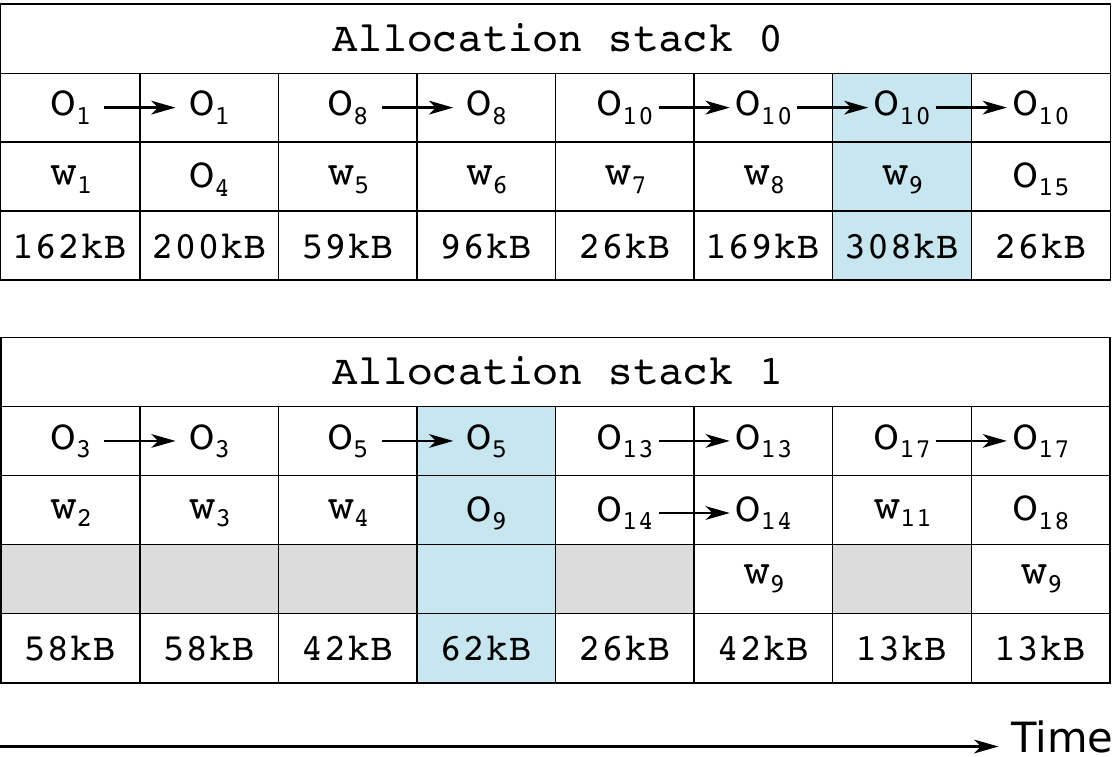}
	\caption{L2 memory allocation sequence.}
	\label{fig:mem_alloc}
\end{figure}

At boot time, before the actual computation loop starts, \textit{i)} we load all the weights, stored in the external flash memory as binary files, in the L3 DRAM memory and \textit{ii)} we call from the fabric controller the \textit{runtime} allocator to reserve two L2 \textit{allocation stacks} (shown in Figure~\ref{fig:mem_alloc}) where intermediate buffers will be allocated and deallocated in a linear fashion.
The choice to use two allocation stacks instead of a single one is because in the latter case, we would need to keep alive up to \SI{665}{\kilo\byte} in L2 due to data dependencies, which is more than the available space.
Our allocation strategy updates the pointer of the next free location in the pre-allocated L2 stack, avoiding the runtime overhead of library allocation/free functions.
We differentiate our strategy between weights and feature maps: for the former, we allocate space just before their related layer and deallocate it just after the layer execution, as also shown in the pseudo-code blocks of Figure~\ref{fig:exec_flow}.
For the latter, due to the residual network bypasses, we often have to prolongate the lifetime of a feature map during the execution of the two following layers (node kernels in Figure~\ref{fig:exec_flow}).
Therefore, for each RES block, there will be an amount of time where three tensors have to be stored at the same time.

Figure~\ref{fig:exec_flow} shows the full execution flow of DroNet related to our solution, annotated with the sequence of node kernels and the L3/L2 memory management blocks.
For the sake of readability, in Figure~\ref{fig:exec_flow}, we report only the first RES block, but this can be generalized also to the others with few minor modifications and updating input, output and weights id.
In the pseudo-code of Figure~\ref{fig:exec_flow}, the second parameter of the \texttt{Alloc} and \texttt{Free} function specifies the allocation buffer (i.e., \texttt{Allocation stack 0} or \texttt{Allocation stack 1} in Figure~\ref{fig:mem_alloc}).
Note that, the $\mu$DMA copies the weights from L3 to L2 just after the destination L2 area is allocated.
The buffers' memory allocation sequence is reported in Figure~\ref{fig:mem_alloc} (from left to right) for the entire DroNet execution.
The columns of the two stacks represent the data needed at each execution step, where $O_i$ and $w_j$ represent the input/output feature maps and weights, respectively.
The last row of each stack reports the total amount of L2 memory required at each step.
Thus, the final dimension of each stack is given by the column with the biggest occupancy (highlighted in light blue in Figure~\ref{fig:mem_alloc}), resulting in \SI{370}{\kilo\byte} of L2 memory.
Therefore, our solution not only allows to the DroNet execution within the L2 memory budget constraint but results in leaving \SI{142}{\kilo\byte} of the L2 still available (i.e., $\sim$28\% of the total) for additional onboard tasks like target following~\cite{palossi_IWASI17}, etc.

%% file: 05-prototype.tex
\section{The PULP-Shield} \label{Sec:prototype}

To host our visual navigation algorithm, we designed a lightweight, modular and configurable printed circuit board (PCB) with highly optimized layout and a form factor compatible with our nano-size quadrotor.
It features a PULP-based GAP8 SoC, two Cypress \textit{HyperBus Memories}\footnote{http://www.cypress.com/products/hyperbus-memory} and an ultra-low power \textit{HiMax} CMOS image sensor\footnote{http://www.himax.com.tw/products/cmos-image-sensor/image-sensors} able to run up to 60~fps with a gray-scale resolution of $320\times240$~pixels with just \SI{4.5}{\milli\watt} of power.
Our pluggable PCB, named \textit{PULP-Shield}, has been designed to be compatible with the \textit{Crazyflie 2.0} (CF) nano-quadrotor\footnote{https://www.bitcraze.io/crazyflie-2\hspace{1cm}}.
This vehicle has been chosen due to its reduced size (i.e., \SI{27}{\gram} of weight and \SI{10}{\centi\meter} of diameter) and its open-source and open-hardware philosophy.
The communication between the PULP chip and the main MCU aboard the nano-drone (i.e., \textit{ST Microelectronics STM32F405}\footnote{http://www.st.com/en/microcontrollers/stm32f405-415.html}) is realized via an SPI interface and two GPIO signals.

\begin{figure}[t]
	\centering
	\includegraphics[width=\columnwidth]{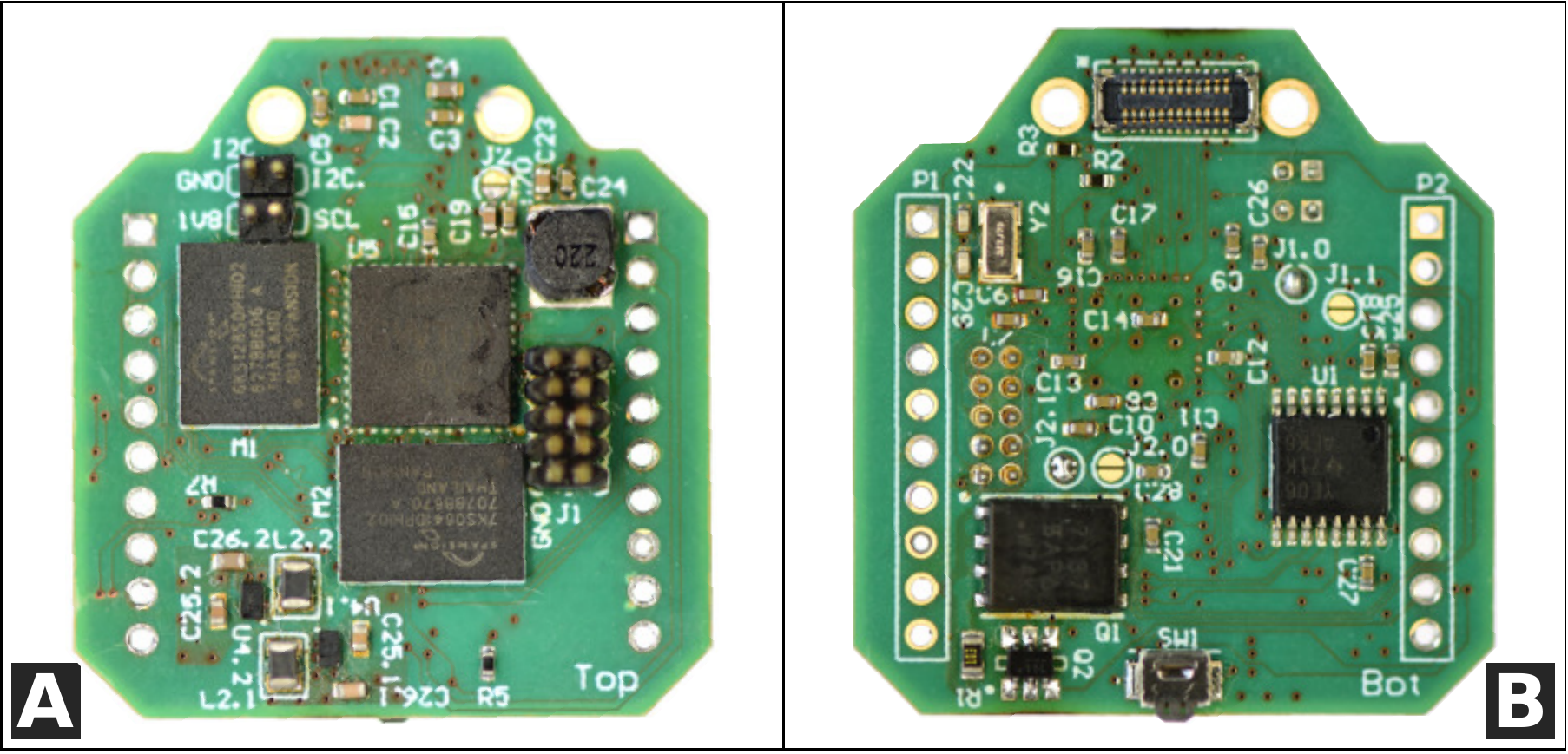}
	\caption{The \textit{PULP-Shield} pluggable PCB. Top view (A) and bottom view (B).}
	\label{fig:pulp-shield-schematic}
\end{figure}

In Figure~\ref{fig:pulp-shield-schematic} the schematic of the proposed PULP-Shield is shown.
Two BGA memory slots allow all possible combinations of \textit{HyperRAM}, \textit{HyperFlash}, and hybrid \textit{HyperFlash/RAM} packages.
In this way, we can select the most appropriate memory configuration given a target application.
We mounted on one slot a \SI{64}{\mega\bit} \textit{HyperRAM} (DRAM) chip and on the other a \SI{128}{\mega\bit} \textit{HyperFlash} memory, embodying the system L3 and the external storage, respectively.

On the PCB (Figure~\ref{fig:pulp-shield-schematic}-B) there is also a camera connector that allows the \textit{HiMax} camera to communicate with the rest of the system through the parallel camera interface (PCI) protocol.
Two mounting holes, on the side of the camera connector, allow plugging a 3D-printed camera holder that can be set either in front-looking or down-looking configuration.
Those two configurations are representative of the most common visual sensors layouts typically embedded in any autonomous flying vehicles.
The front-looking configuration can be used for many navigation tasks like path planning~\cite{iot_PathPlanning2}, obstacle avoidance~\cite{iot_PathPlanning1}, trajectory optimization~\cite{7989679}, to name a few.
Instead, the down-looking camera configuration is usually chosen for stabilization tasks like distance estimation~\cite{palossiDATE17}, way-point tracking, and positioning~\cite{8088164}, etc. 

On the shield, there are also a JTAG connector for debug purposes and an external I2C plug for future development.
Two headers, located on both sides of the PCB, grant a steady physical connection with the drone and at the same time, they bring the shield power supply and allow communication with the CF through the GPIOs and the SPI interface.
The form factor of our final PULP-Shield prototype is 30$\times$\SI{28}{\milli\meter}, and it weighs $\sim$\SI{5}{\gram} (including all components), well below the payload limit imposed by the nano-quadcopter.

\begin{figure}[h]
	\centering
	\includegraphics[width=\columnwidth]{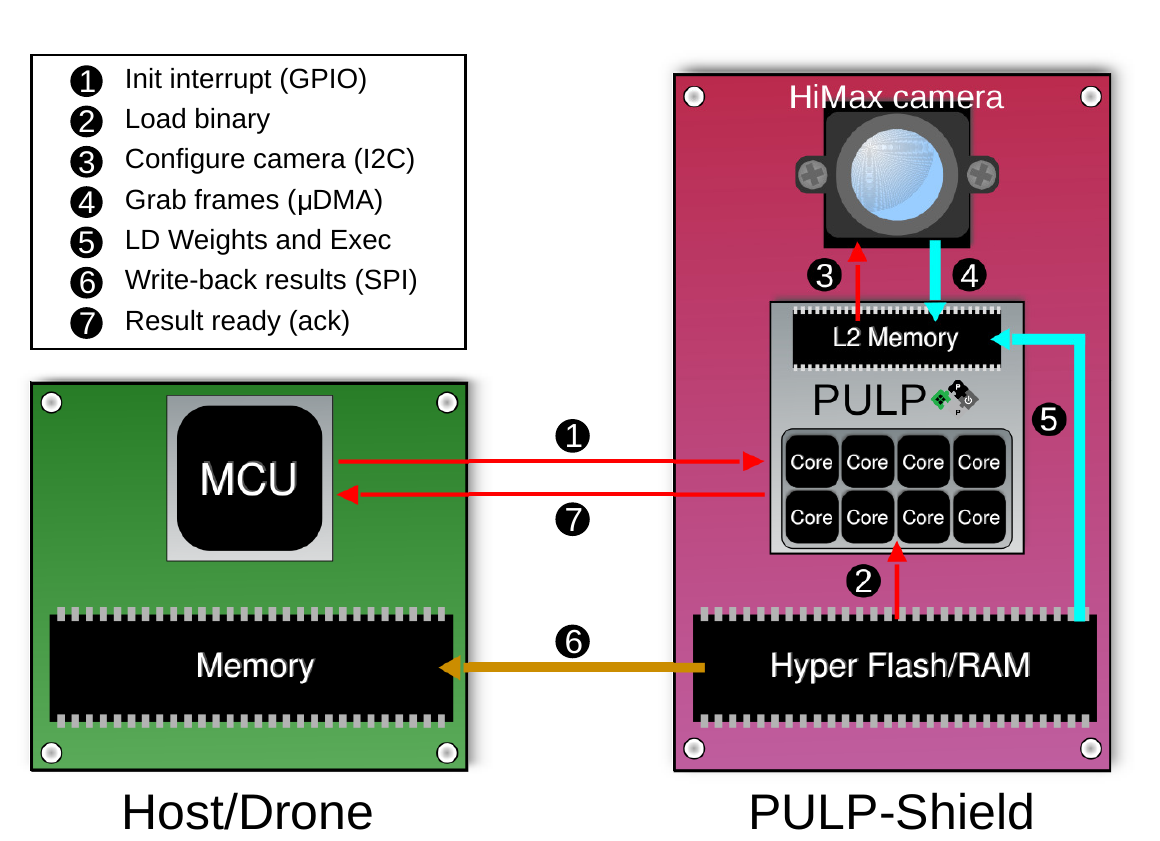}
	\caption{Example of interaction between the \textit{PULP-Shield} and the drone.}
	\label{fig:pulp-shield-model}
\end{figure}

Similarly to what has been presented in \cite{contiDATE16}, the PULP-Shield embodies the \textit{Host-Accelerator} architectural paradigm, where the CF's MCU offloads the intensive visual navigation workload to the PULP accelerator.
As depicted in Figure~\ref{fig:pulp-shield-model} the interaction starts from the host, which wakes up the accelerator with a GPIO interrupt \circled{1}.
Then, the accelerator fetches from its external \textit{HyperFlash} storage the kernel (stored as a binary file) to be executed: DroNet in our case \circled{2}.
Note that, in this first part of the protocol the host can also specify which kernel should be executed, as well as a sequence of several pre-loaded ones available on the external Flash storage.
At this point, the GAP8 SoC can configure the \textit{HiMax} camera via an internal I2C \circled{3} and start to transfer the frames from the sensor to the L2 shared memory through the $\mu$DMA \circled{4}. 
All additional data, like the weights used in our CNN, can be loaded from the DRAM/Flash memory and parallel execution is started on the accelerator \circled{5}.
Lastly, the results of the computation are returned to the drone's MCU via SPI \circled{6}, and the same host is acknowledged about the available results with a final interrupt over GPIO \circled{7}.
Note that, the transfer of a new frame is performed by the $\mu$DMA overlapping the CNN computation on the previous frame performed in the \textsc{cluster}.

Even if the PULP-Shield has been developed specifically to fit the CF quadcopter, its basic concept and the functionality it provides are quite general, and portable to any drone based on an SPI-equipped MCU and more generally to a generic IoT node requiring visual processing capabilities. 
The system-level architectural template it is based on is meant for minimizing data transfers (i.e., exploiting locality of data) and communication overhead between the main MCU and the accelerator -- without depending on the internal microarchitecture of either one.

%% file: 06-results.tex
\section{Experimental Results} \label{Sec:results}

\begin{figure*}[t]
	\centering
	\includegraphics[width=\textwidth]{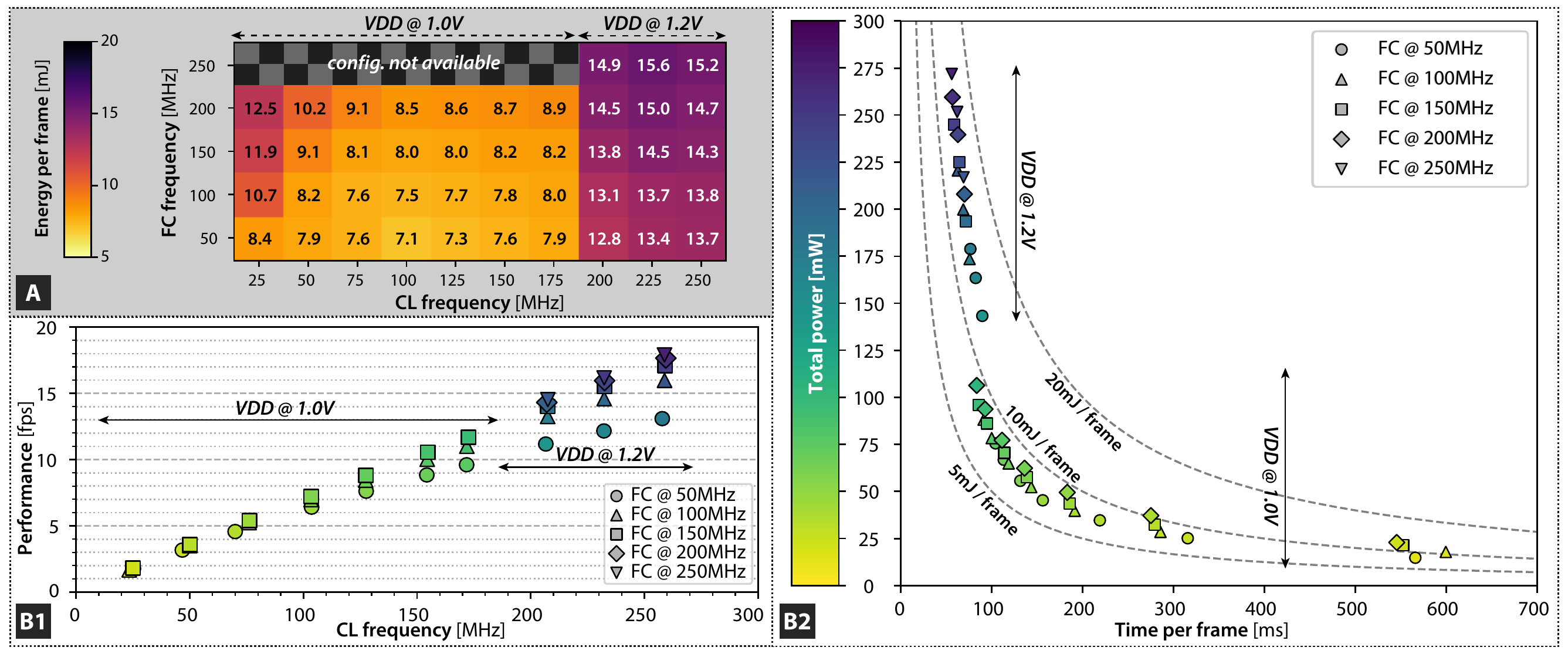}
	\caption{A) Heat map showing the energy per frame in all tested configurations of GAP8 with  VDD@\efficientVDD{} and  VDD@\fastVDD{}; B1) DroNet performance in frames per second (fps) in all tested configurations (coloring is proportional to total system power); B2) DroNet total system power vs. time per frame in all tested configurations; dashed gray lines show the levels of energy efficiency in mJ/frame.}
	\label{fig:op}
\end{figure*}

In this section we present the experimental evaluation of our visual navigation engine, considering three primary metrics: \textit{i)} the capability of respecting a given real-time deadline, \textit{ii)} the ability of performing all the required computations within the allowed power budget and \textit{iii)} the final accuracy of the closed-loop control, given as reaction time w.r.t. an unexpected obstacle.
All the results are based on the PULP-Shield configuration presented in Section~\ref{Sec:prototype}.

\subsection{Performance \& Power Consumption} 

We measured wall-time performance and power consumption by sweeping between several operating modes on GAP8.
We focused on operating at the lowest (\SI{1.0}{\volt}) and highest (\SI{1.2}{\volt}) supported core VDD voltages.
We swept the operating frequency between 50 and \SI{250}{\mega\hertz}, well beyond the GAP8 officially supported configuration\footnote{https://greenwaves-technologies.com/gap8-datasheet}.
Figure~\ref{fig:op} provides a complete view of the power consumption in all experimentally possible operating modes of GAP8 on the DroNet application while sweeping both \textsc{Fabric Ctrl} (FC) and \textsc{Cluster} (CL) clock frequency, both at \SI{1.0}{\volt} and \SI{1.2}{\volt} and the related achievable frame-rate.
Figure~\ref{fig:op}-A shows the energy-efficiency of all available configurations as a heat map, where VDD@\efficientVDD{}, FC@\efficientFCfmax{}, and CL@\efficientCLfmax{} represent the most energy efficient one.
In Figure~\ref{fig:op}-B1 we report performance as frame-rate and total power consumption measured before the internal DC/DC converter utilized on the SoC. 
Selecting a VDD operating point of \fastVDD{} would increase both power and performance up to \fastpoweravg{} and \fastfps{}.
We found the SoC to be working correctly @ \SI{1.0}{\volt} for frequencies up to $\sim$\SI{175}{\mega\hertz}; we note that as expected when operating @ \SI{1.0}{\volt} there is a definite advantage in terms of energy efficiency.
Therefore, for the sake of readability, in Figure~\ref{fig:op} we avoid showing configurations of VDD \SI{1.2}{\volt} that would reach the same performance of VDD \SI{1.0}{\volt} at a higher cost in term of power.
Similarly, in Figure~\ref{fig:op}-B2 we report power consumption vs time to compute one frame.

\begin{table}[h]
\renewcommand{\arraystretch}{1.3}
\caption{Power consumption \& Execution time per frame of \textit{DroNet} on GAP8 VDD@\efficientVDD{}, FC@\efficientFCfmax{}, CL@\efficientCLfmax{}.}
\label{tab:power_time}
\centering
\scriptsize
\begin{tabular}{|c|c|c|c|}
\hline
Layer & AVG Power [mW] & Exec Time [ms] & L3-L2 Time [ms] \\
\hline
conv\_1 + pool  & 47.1 & 22.6 & 0.1\\ 
\hline
ReLU            & 24.8 & 0.9 & ---\\ 
\hline
conv\_2 + ReLU  & 38.7 & 17.3 & 0.6\\ 
\hline
conv\_3         & 38.0 & 14.3 & 0.6\\ 
\hline
conv\_4         & 43.6 & 7.3 & 0.1\\ 
\hline
add             & 38.9 & 0.3 & ---\\
\hline
ReLU            & 27.6 & 0.2 & ---\\
\hline
conv\_5 + ReLU  & 37.7 & 9.3 & 1.2\\ 
\hline
conv\_6         & 34.8 & 17.0 & 2.4\\
\hline
conv\_7         & 32.7 & 4.2 & 0.2\\ 
\hline
add             & 24.3 & 0.3 & ---\\
\hline
ReLU            & 20.5 & 0.3 & ---\\
\hline
conv\_8 + ReLU  & 33.1 & 13.0 & 4.7\\ 
\hline
conv\_9         & 31.9 & 24.8 & 9.4\\ 
\hline
conv\_10        & 41.9 & 5.4 & 0.5\\ 
\hline
add + ReLU      & 24.4 & 0.3 & ---\\
\hline
fully\_1        & 13.0 & 0.1 & 0.4\\ 
\hline
fully\_2        & 13.0 & 0.1 & 0.4\\ 
\hline
\end{tabular}
\end{table}

\begin{figure*}[t]
	\centering
	\includegraphics[width=\textwidth]{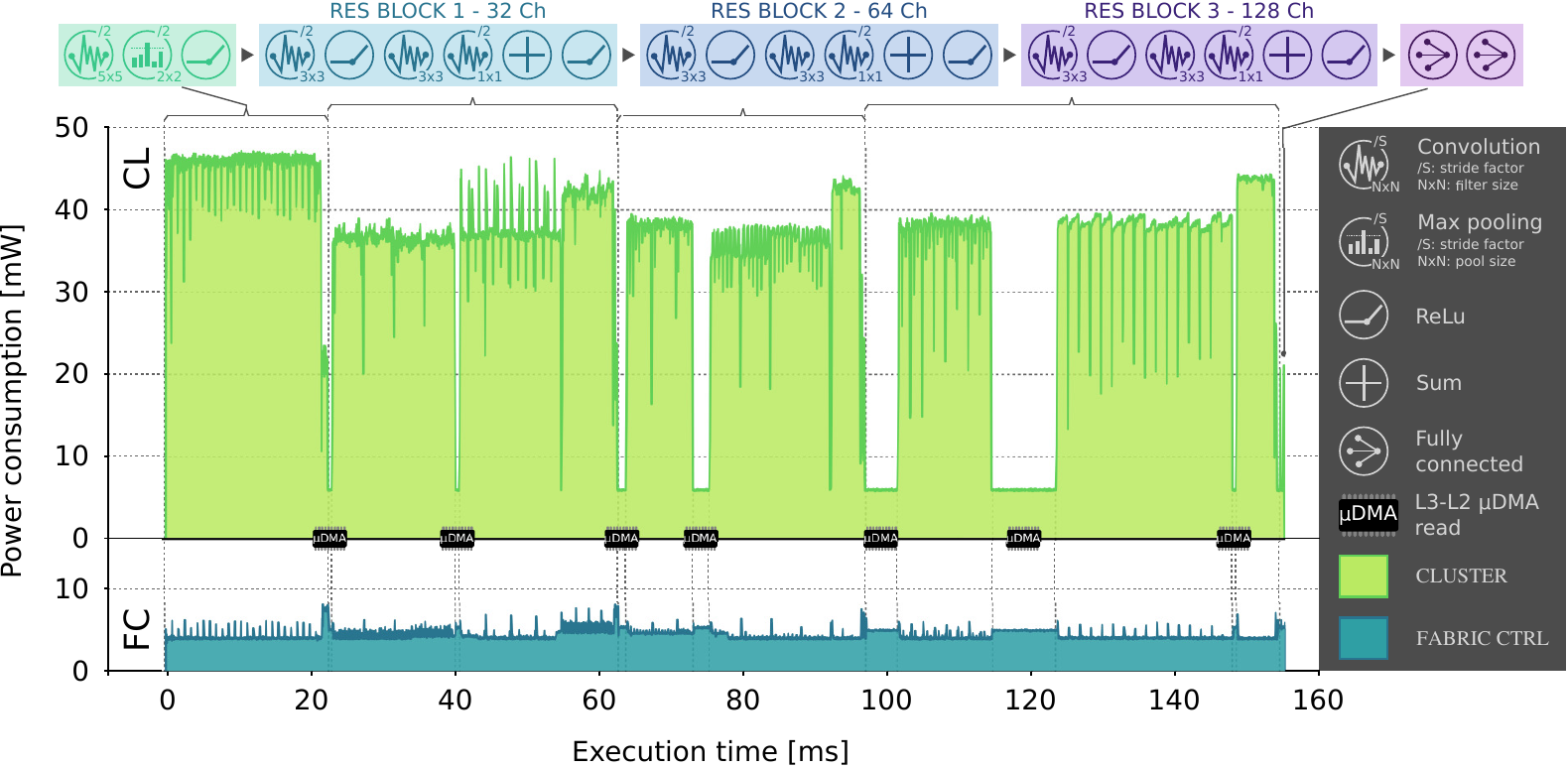}
	\caption{Power traces per layer of \textit{DroNet}, measured at VDD@\efficientVDD{}, FC@\efficientFCfmax{}, CL@\efficientCLfmax{}, the symbols on top of the plot indicate the computation stage associated with each visible phase in the power trace. Measurements are taken after internal DC/DC converter (i.e., accounting for both \textsc{fabric ctrl} and \textsc{cluster}).
	}
	\label{fig:power-layer}
\end{figure*}

In Figure~\ref{fig:power-layer} we present the power traces for full end-to-end execution of DroNet, measured using a bench DC power analyzer\footnote{www.keysight.com/en/pd-1842303-pn-N6705B}.
The power traces are measured by powering the GAP8 SoC, with the most energy-efficient configuration at \efficientVDD{} core voltage and operating at \efficientFCfmax{} on FC and \efficientCLfmax{} on CL.
The detailed average power consumption (including both the \textsc{FC} and \textsc{CL} domains) is reported in Table~\ref{tab:power_time}.
The peak power consumption of \efficientlayerpowermax{} is associated to the $1^{st}$ convolutional layer; we used this value to compute the overall power envelope of our node.
Instead, the minimum power consumption is given by the two last fully connected layers consuming \efficientlayerpowermin{} each.
The average power consumption, weighted throughout each layer, is \efficientlayerpoweravg{}, which grows to \efficientpoweravg{} including also the losses on the internal DC/DC converter (not included in Figure~\ref{fig:power-layer}).
In the full DroNet execution, layers are interposed with L3-L2 data transfers, happening with the CL cores in a clock-gated state, which accounts for $\sim$7\% of the overall execution time.
Therefore, power consumption for the entire board settles to \efficientpowerboard{} if we also consider the cost of L3 memory access and the onboard ULP camera.

\begin{figure}[h]
	\centering
	\includegraphics[width=\columnwidth]{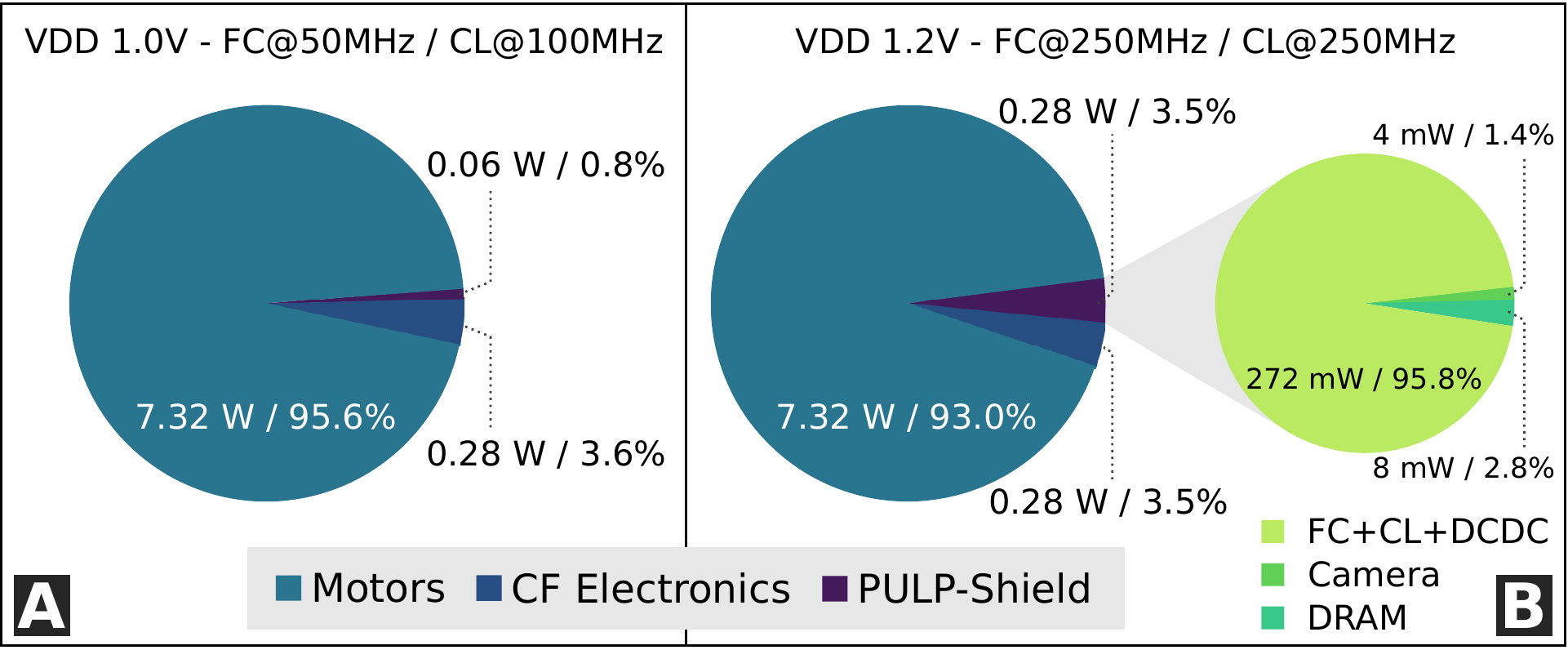}
	\caption{Power envelope break-down of the entire cyber-physical system running at FC@\efficientFCfmax{}-CL@\efficientCLfmax{} (A) and FC@\fastFCfmax{}-CL@\fastCLfmax{} (B) with \textit{PULP-Shield} zoom-in (on the right).}
	\label{fig:break_down}
\end{figure}

In Figure~\ref{fig:break_down} is reported the power break-down for the complete cyber-physical system and proposed PULP-Shield. 
Our nano-quadcopter is equipped with a \SI{240}{\milli\ampere\hour} \SI{3.7}{\volt} LiPo battery enabling a flight time of 7 minutes under standard conditions, which results in average power consumption of \SI{7.6}{\watt}.
The power consumption of all the electronics aboard the original drone amounts to \SI{277}{\milli\watt} leaving $\sim$\SI{7.3}{\watt} for the four rotors.
The electronics consumption is given by the 2 MCUs included in the quadrotor and all the additional devices (e.g., sensors, LEDs, etc.).
In addition to that, introducing the PULP-Shield, we increase the peak power envelope by \efficientpowerboard{} using the most energy-efficient configuration and by \fastpowerboard{} selecting the fastest setting (\efficientpowerpercent{} and \fastpowerpercent{} of the total, respectively).
On the PULP-Shield we consider the HyperRAM is operating at full speed only for the time required for L3-L2 data transfers (as shown in Table~\ref{tab:power_time}) with an average power consumption of \SI{8}{\milli\watt} for the fastest configuration, as reported in Figure~\ref{fig:break_down}-B.
Notice that this is a worst-case figure, taking account of both the GAP8 SoC and the HyperRAM operating at full speed simultaneously.
The power break-down of our visual navigation module can be seen on the right of Figure~\ref{fig:break_down}-B, where we include the computational unit, the L3 external DRAM memory, and the ultra-low power camera.
As onboard computation accounts for roughly 5\% of the overall power consumption (propellers, sensors, computation and control, \textit{cfr} Section~\ref{Sec:introduction}), our PULP-Shield enables the execution of the DroNet network (and potentially more) in all configurations within the given power envelope.

\begin{table}[h]
\renewcommand{\arraystretch}{1.3}
\caption{\textit{CrazyFlie} (CF) lifetime with and without \textit{PULP-Shield} (both turned off and running \textit{DroNet} at VDD@\efficientVDD{}, FC@\efficientFCfmax{}, CL@\efficientCLfmax{}).}
\label{tab:lifetime}
\centering
\scriptsize
\begin{tabular}{|c|c|c|c|c|}
\hline
& Original CF & CF + \textit{PULP-Shield} (off) & CF + \textit{PULP-Shield} (on) \\
\hline
Lifetime & $\sim$\SI{440}{\second} & $\sim$\SI{350}{\second} & $\sim$\SI{340}{\second} \\
\hline
\end{tabular}
\end{table}

Finally, in our last experiment, we evaluate the cost in terms of operating lifetime of carrying the physical payload of the PULP-Shield and of executing the DroNet workload.
To ensure a fair measurement, we decoupled the DroNet output from the nano-drone control and statically set it to \textit{hover} (i.e., keep constant hight over time) at \SI{0.5}{\meter} from the ground.
We targeted three different configurations: \textit{i}) the original \textit{CrazyFlie} (CF) without any PULP-Shield; \textit{ii}) PULP-Shield plugged but never turned on, to evaluate the lifetime reduction due to the additional weight introduced; \textit{iii}) PULP-Shield turned on executing DroNet at VDD@\efficientVDD{}, FC@\efficientFCfmax{}, CL@\efficientCLfmax{}.
Our results are summarized in Table~\ref{tab:lifetime} and as expected the biggest reduction in the lifetime is given by the increased weight.
The flight time of the original nano-drone, with one battery fully charged, is $\sim$\mbox{\SI{440}{\second}}.
This lifetime drops to $\sim$\mbox{\SI{350}{\second}} when the drone is carrying the PULP-Shield (turned off) and to $\sim$\mbox{\SI{340}{\second}} when the shield is executing DroNet.
Ultimately, the price for our visual navigation engine is $\sim$22\% of the original lifetime.

\subsection{State-of-the-Art Comparison \& Discussion} \label{Sec:discussion}

To compare and validate our experimental results with respect to the current state-of-the-art, we targeted the most efficient CNN implementation currently available for microcontrollers, namely CMSIS-NN~\cite{LaiCMSISNNEfficientNeural2018}.
At peak performance in a synthetic test, this fully optimized library can achieve as much as 0.69\,MAC/cycle on convolutions, operating on \textit{Fixed8} data that is internally converted to \textit{Fixed16} in the inner loop.

By contrast, we operate directly on \textit{Fixed16} and achieve a peak performance of 0.64~MAC/cycle/core in a similar scenario (on the $6^{th}$ layer of DroNet, 3$\times$3 convolution).
The bypasses and the final layers are a bit less efficient, yielding an overall weighted peak throughput of 0.53~MAC/cycle/core on convolutional layers, which constitute the vast majority of the execution time.

\begin{table}[h]
\renewcommand{\arraystretch}{1.3}
\caption{\textsc{Cluster}-cycle break-down for processing one frame on the GAP8 both FC and CL @ \SI{50}{\mega\hertz}.}
\label{tab:cycle_breakdown}
\centering
\scriptsize
\begin{tabular}{|c|c|c|c|c|}
\hline
& $\mu DMA$ L3/L2 & $DMA$ L2/L1 & Computation & Total\\
\hline
Cycles & \SI{1.03}{M} & \SI{0.11}{M} & \SI{13.47}{M} & \SI{14.61}{M} \\
\hline
\end{tabular}
\end{table}

\begin{figure*}[t]
	\centering
	\includegraphics[width=\textwidth]{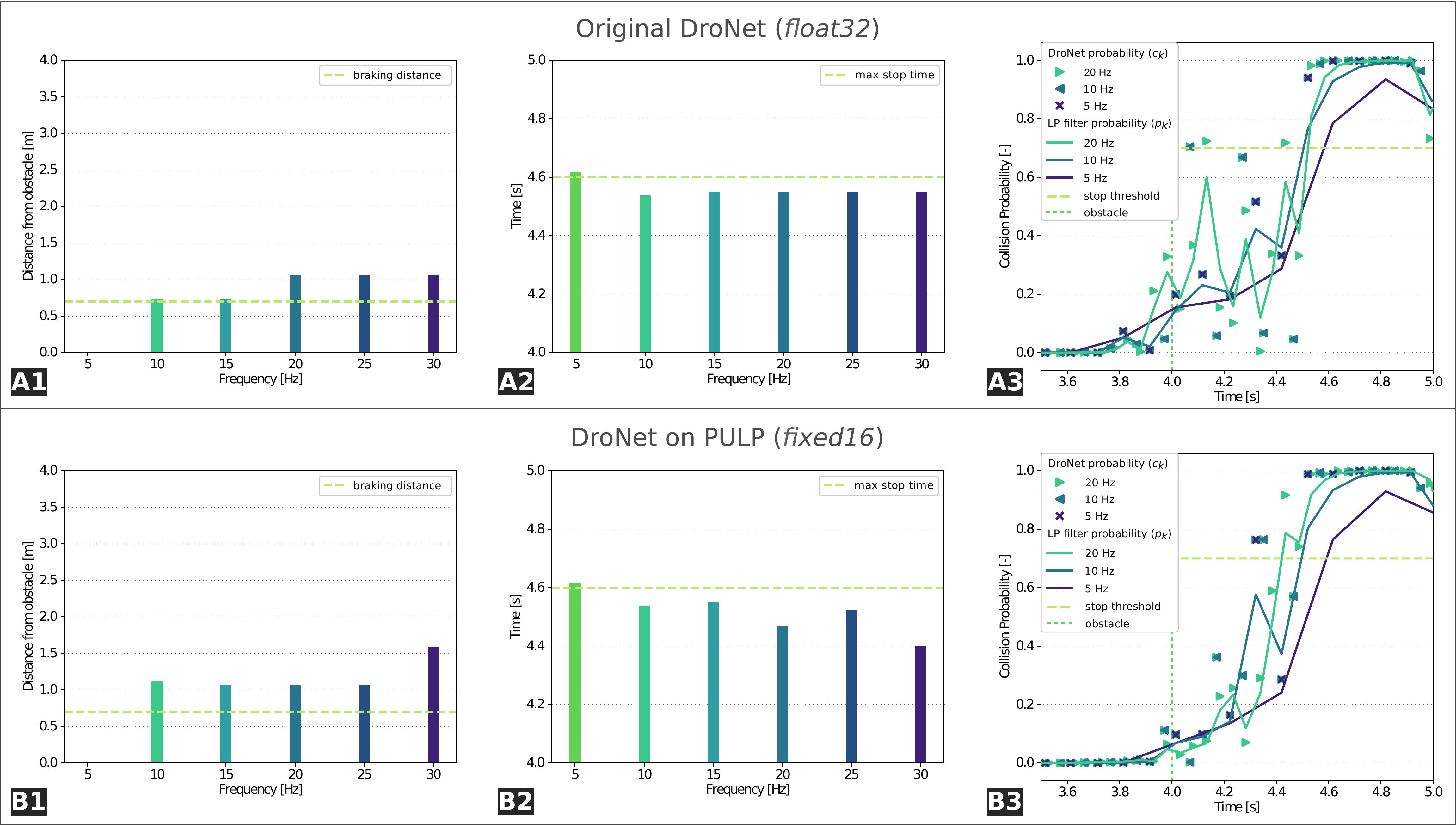}
	\caption{Performance comparison between the original (A1-3) and quantized (B1-3) \textit{DroNet} architectures. Stop command time (A1 and B1), minimum distance from the obstacle (A2 and B2) and collision probability as the output of both CNN and low-pass filter (A3 and B3).}
	\label{fig:control_acc_x6}
\end{figure*}

In Table \ref{tab:cycle_breakdown} we report the execution breakdown per frame for all activities performed by our CNN.
We can see how the L3-L2 transfers (not overlapped to computation) and the non-overlapping part of L2-L1 transfers account for $\sim$1.14~Mcycles of the overall execution time.
Then, considering $\sim$41~MMAC for the original CNN, in the ideal peak-throughput case of 4.28~MAC/cycle we would need $\sim$10~Mcycles for computing one frame, instead of our measured 13.47~Mcycles.
The overhead is due to inevitable non-idealities such as sub-optimal load balancing in layers exposing limited spatial parallelism as well as tiling control loops and the marshaling stage required by padded convolutions.
Considering all of the effects mentioned above (i.e., computation non-idealities as well as memory transfers), we achieve a real throughput of 2.81~MAC/cycle in the DroNet execution -- still 4$\times$ better than the CMSIS-NN peak performance.

To further concretize the comparison, we take as an example target a top-notch high-performance microcontroller: an STM32H7\footnote{http://www.st.com/en/microcontrollers/stm32h7-series.html} sporting a Cortex-M7 core and capable of operating at up to \SI{400}{\mega\hertz}.
Without considering any data movement overhead, and taking into account only peak performance, this would be able to achieve up to 276~MMAC/s @ \SI{346}{\milli\watt}.
By comparison, our system can achieve an average performance of 281~MMAC/s with the most power-efficient configuration @ \efficientpoweravg{}, i.e. same performance within a 5.4$\times$ smaller power budget.
Moreover, if we consider our peak-throughput configuration (where both FC and CL are running @ \SI{250}{\mega\hertz}) we can deliver up to 702~MMAC/s @ \SI{272}{\milli\watt}: 2.5$\times$ better with 21\% less power. 
Even if it were possible to linearly up-scale the performance of this microcontroller to the same level of our system, it would consume $\sim$\SI{880}{\milli\watt}, which would constitute largely more than the 5\% of power envelope typically dedicated to onboard computation on nano-UAV systems \cite{Wood2017}.
This confirms that the parallel-ultra-low power approach adopted in our visual navigation engine significantly outperforms sequential processing in terms of energy efficiency, without compromising programmability and flexibility.

\subsection{Control Accuracy} 

To fully exploit the natural inertial agility of a lightweight nano-quadrotor as the \textit{Crazyflie 2.0} used in our prototype, fast onboard perception is required. 
To evaluate the agility of our integrated system, we perform an experiment in which our flying platform is required to react to a sudden obstacle occluding its way.
With this experiment, we aim to demonstrate that the PULP-Shield computational resources are enough to make full use of the platform agility.
As mentioned in Section~\ref{Sec:deployment}, for the final deployment of DroNet on the PULP-Shield, we select the network trained with \texttt{Fixed16} quantization, $2\times2$ max-pool receptive field, and fine-tuning dataset.
The choice is justified by both the quantization requirement of the GAP8 SoC and the model performance, superior to other viable alternatives (see Table~\ref{tab:dronet_pulp_accuracy}).

The experimental setting is as follows: we collect a dataset of images by manually flying the drone over a straight path of \SI{20}{\meter} at an average speed of \SI{4}{\meter/\second}. 
At the beginning of the test, the path is entirely free from obstacles. 
At $T=$\SI{4}{\second} after the start of the experiment, an obstacle appears at the end of the track, leaving \SI{4}{\meter} free for breaking and stopping.
The system is then required to raise a stop signal soon enough to avoid the collision.
As we show in the additional video, our integrated system can control the nano-drone in closed-loop. 
However, for safety reasons and to avoid damaging the platform, we don't control the nano-drone in closed-loop during this experiment. 
Instead, we process the frames collected with manual flight offline.
The collected dataset is used to study the relation between the system operational frequencies and the drone's reaction time.

As in the original implementation of~\cite{dronet}, network's predictions are low-pass filtered to decrease high-frequency noise.
In detail, the collision probability $p_k$ is a low-pass filtered version of the raw network output $c_k$ ($\alpha=0.7$):

\begin{equation}
        p_k = (1 - \alpha) p_{k-1} + \alpha c_k,                                                    
\end{equation}

Figure~\ref{fig:control_acc_x6} (A3-B3) illustrates the predicted collision probability of the original and quantized DroNet CNN as a function of time.
In the plots, we show both $c_k$ and $p_k$ at different frequencies, the former reported as markers, whereas the latter is shown as a continuous line.
A horizontal dashed orange line shows the threshold for sending a stop signal to the control loop ($p_k > 0.7$), and a vertical red dashed line highlights the time at which the obstacle becomes visible ($T=$\SI{4}{\second}). 

To quantitatively evaluate the performance of our system at different operational frequencies, we computed the maximum time and the minimum distance from the object at which the stop command should be given to avoid the collision.
We deployed the \textit{Crazyflie 2.0} parameters from~\cite{forster2015system} and the classical quadrotor motion model from~\cite{Mahony_2012} to \textit{analytically} compute those two quantities.
From this analysis, we derived a minimum stopping time of \SI{400}{\milli\second} and a braking distance of \SI{0.7}{\meter}, assuming the platform moves with a speed of \SI{4}{\meter/\second} when it detects the obstacle. 

In Figure~\ref{fig:control_acc_x6} (A1-2, B1-2) we illustrate a performance comparison between our quantized system and the original implementation of~\cite{dronet}.
Despite quantization, our network outperforms~\cite{dronet} in term of collision detection, and can react more quickly to sudden obstacles even at low operational frequencies.
This is in accordance with the results of Table~\ref{tab:dronet_pulp_accuracy}, and mainly due to the fine-tuning of our network to the HiMax camera images.

Both the quantized and original architecture share however a similar behaviour at different operational frequencies.
More specifically, both fail to detect obstacles at very low frequencies (i.e., \SI{5}{\hertz}), but successfully avoid the collision at higher rates.
Interestingly, increasing the system frequencies does not always improve performance; it can be observed in Figure~\ref{fig:control_acc_x6}-B2, where performance at \SI{20}{\hertz} is better than at \SI{25}{\hertz}. 
From Figure~\ref{fig:control_acc_x6} we can observe that inference at \SI{10}{\hertz} allows the drone to brake in time and avoid the collision. 
This confirms that our system, processing up to \fastfps{}, can \textit{i)} make use of the agility of the \textit{Crazyflie 2.0} and \textit{ii)} be deployed in the same way as the original method to navigate in indoor/outdoor environments while avoiding dynamic obstacles.
A video showing the performance of the system controlled in closed-loop can be seen at the following link: \videolink.

%% file: 07-conclusion.tex
\section{Conclusion} \label{Sec:conclusion}

Nano- and pico-sized UAVs are ideal IoT nodes; due to their size and physical footprint, they can act as mobile IoT hubs, smart sensors and data collectors for tasks such as surveillance, inspection, etc.
However, to be able to perform these tasks, they must be capable of autonomous navigation of environments such as urban streets, industrial facilities and other hazardous or otherwise challenging areas.
In this work, we present a complete deployment methodology targeted at enabling execution of complex deep learning algorithms directly aboard resource-constrained milliwatt-scale nodes. 
We provide the first (to the best of our knowledge) completely vertically integrated hardware/software visual navigation engine for autonomous nano-UAVs with completely onboard computation -- and thus potentially able to operate in conditions in which the latency or the additional power cost of a wirelessly-connected centralized solution.

Our system, based on a \textit{GreenWaves Technologies} GAP8 SoC used as an accelerator coupled with the STM32 MCU on the \textit{CrazyFlie 2.0} nano-UAV, supports real-time computation of DroNet, an advanced CNN-based autonomous navigation algorithm.
Experimental results show a performance of \efficientfps{} @ \efficientpowerboard{} selecting the most energy-efficient SoC configuration, that can scale up to \fastfps{} within an average power budget for computation of \fastpowerboard{}.
This is achieved without quality-of-results loss with respect to the baseline system on which DroNet was deployed: a COTS standard-size UAV connected with a remote PC, on which the CNN was running at \SI{20}{fps}.
Our results show that both systems can detect obstacles fast enough to be able to safely fly at high speed, \SI{4}{\meter/\second} in the case of the \textit{CrazyFlie 2.0}.
To further paving the way for a vast number of advanced use-cases of autonomous nano-UAVs as IoT-connected mobile smart sensors, we release open-source our PULP-Shield design and all code running on it, as well as datasets and trained networks.